%% file: 00_main.tex
\documentclass[letterpaper]{article}
\usepackage[margin=1in,footskip=0.25in]{geometry}

\usepackage{graphicx}
\usepackage{microtype}                 
\PassOptionsToPackage{warn}{textcomp}  
\usepackage{textcomp}                  
\usepackage{mathptmx}                  
\usepackage{times}                     
\usepackage{cite}                      
\usepackage{tabu}                      
\usepackage{booktabs}                  
\usepackage{xspace}
\usepackage{wrapfig}
\usepackage{multirow}
\usepackage{subcaption}
\usepackage{balance}
\usepackage{enumitem}
\usepackage{xcolor}
\usepackage{url}
\usepackage[margin=3cm]{caption}

\newcommand{\lvo}{LAMVI-1\xspace}
\newcommand{\lvt}{LAMVI-2\xspace}
\newcommand{\wevi}{WEVI\xspace}

\begin{document}

\title{LAMVI-2: A Visual Tool for Comparing and Tuning \\ Word Embedding Models}

\author{Xin Rong, Joshua Luckson, and Eytan Adar}

\maketitle


\abstract{Tuning machine learning models, particularly deep learning architectures, is a complex process. Automated hyperparameter tuning algorithms often depend on specific optimization metrics. However, in many situations, a developer trades one metric against another: accuracy versus overfitting, precision versus recall, smaller models and accuracy, etc.  With deep learning, not only are the model's representations opaque, the model's behavior when parameters ``knobs'' are changed may also be unpredictable. Thus, picking the ``best'' model often requires time-consuming model comparison. In this work, we introduce \lvt, a visual analytics system to support a developer in comparing hyperparameter settings and outcomes. By focusing on word-embedding models (``deep learning for text'') we integrate views to compare both high-level statistics as well as internal model behaviors (e.g., comparing word `distances'). We demonstrate how developers can work with \lvt to more quickly and accurately narrow down an appropriate and effective application-specific model.
}

\input{01_intro}
\input{02_related}
\input{03_design}
\input{04_system}

\input{05_casestudy}

\input{06_eval}

\input{07_discussion}

\input{08_conclusion}

\input{99_ack}

\bibliographystyle{abbrv-doi-narrow}

\bibliography{template}
\end{document}

%% file: 01_intro.tex
\section{Introduction}

Hyperparameter tuning in modern machine learning systems is critical for building effective models. However, the process--in particular for deep learning architectures--is often unintuitive and time consuming. Due to their effectiveness and wide availability, deep learning architectures are nonetheless popular as developers are willing to struggle with the tuning process. Standardized platforms (e.g., Tensorflow, Keras, PyTorch, etc.) have allowed developers to easily release their models through open-source platforms such as GitXiv and Github. The lack of clarity on which settings to use (and where and when) presents an added challenge to `model consumers.' These down-stream developers must contend with finding ways to \textit{retune} or \textit{fine-tune} models to extend, retrain, or perform new evaluation with new data.

Take for example word embedding models such as word2vec~\cite{mikolov2007language}, GloVe~\cite{pennington2014glove}, or Doc2Vec~\cite{le2014distributed}. These models capture the syntactic (and possibly semantic) relationships between terms by embedding words in a high-dimensional vector space by learning the weights for each vector dimension. The magnitude and direction of the vectors \textit{between} can be used to find related terms (nearest neighbors). Additionally, that can be used for other semantic labeling tasks, such as identifying synonyms or other relationships (e.g., hypernyms, antonyms, etc.). Embedding models, unfortunately, are not always easy to train. Even relatively simple ones, such as word2vec, easily have a dozen hyperparameters that control model construction. Everything from window size (how much text to consider around terms) to learning rate (alpha) to vector size (in what dimensional space) to architecture type (`cbow' or 'skip-gram') to sampling types can be modified. It is difficult to understand which hyperparameter 'knobs' to adjust (and how) in order to achieve a good outcome. Additionally, publicly released models are often trained and evaluated on standard corpora (e.g., Wikipedia and SEMEVAL~\cite{semeval}). Real-world use on new corpora requires identifying both the right ground-truth and the right hyperparameters for that corpora.

\begin{figure}
  \centering
  \includegraphics[width=\linewidth]{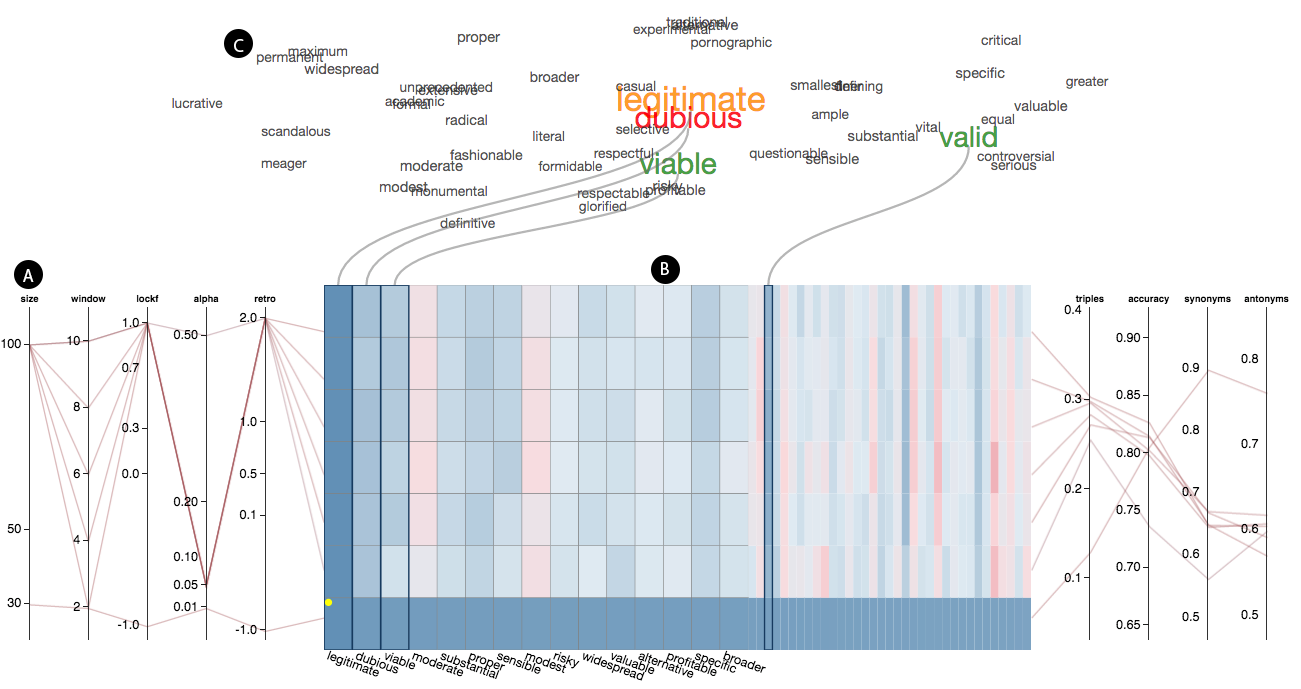}
  \caption{Screenshot of \lvt. A parallel coordinates visualization (A) provides access to models tuned with different hyperparameters (left) and evaluated on different metrics (right). The central heatmap supports model inspection (rows). In this case we see the distances of various terms (columns) to the query word ``legitimate'' encoded through color. A linked t-SNE visualization (B) provides a dimension-reduced view of the embedding. The user can interact with the visualization to understand the trade-offs between hyperparameters and metric, isolate models that perform well, identify outliers (models and instances), and provide examples for evaluation/supervision.}
	\label{fig:lamvi2-screenshot}
\end{figure}

Automated hyperparameter optimization is one common solution.  The simplest variant, grid-search, progressively varies each parameter to find a good model. More complex schemes use other search mechanisms~\cite{ bergstra2012random, NIPS2011_4443,  greff2016lstm, NIPS2012_4522}. An optimizer will attempt to identify the best hyperparameter set given a specific criteria (e.g., to get the best accuracy, use this configuration). Optimization suffers from two problems. First, each `run' may be expensive so constraining the hyper-parameter search is often useful (e.g., identifying a range). Second, not all real world problems have a single optimization metric and many require trade-offs. For example, one may trade-off precision and recall, or increased model size with increased accuracy, but at the cost of space resources (important on devices such as cell phones). In some situations a generic model may be constructed with multiple uses in mind, and maximum performance across all applications can not be simultaneously achieved. Finally, optimization algorithms often assume `stable' test data. A common scenario is that in adapting a model to a new scenario, the developer will find problems in the training or test data and will add data or labels as part of tuning.

Because of these limitations and, debugging and tuning models often requires human interaction. To support human-in-the-loop model tuning, we introduce \lvt (LAnguage Model Visual Inspector 2.0). \lvt is a visualization-driven tool specifically focused on debugging neural embedding models (Figure~\ref{fig:lamvi2-screenshot}). In implementing a variety of prototypes and through task analysis, we identified key decision points in model construction and tuning. Rather than a broad solution, \lvt uses visual channels to target key features of the data/model that are useful in making decisions. The system allows the developer to focus on hyperparameter interaction and to identify how these parameters impact the model at both a fine level (e.g., specific term embeddings and distances) and in more broad measures of model performance. Through \lvt, the developer can construct new model instances and add these to the comparison set. Through the interface, developers can find the best hyperparameters for their model by identifying complex correlations and interactions (e.g., increasing parameter X improves metric Y, but only up to the value Z). Outliers, that indicate possible problems in training or evaluation data, are also apparent in the interface. By integrating a labeling system directly into the interface, \lvt users can also act to `correct' or `extend' model training and evaluation based on their observations.

While previous solutions, include our own work, focused on inspecting \textit{single} trained models \lvt is specifically constructed for \textit{multiple}-model scenarios use cases. This is achieved by combining multiple visual encodings and interaction patterns for tuning-related tasks. As can be seen in Figure~\ref{fig:lamvi2-screenshot}, we 'anchor' \lvt on a parallel coordinates style visualization (showing parameters and output metrics) and link it to heatmaps (capturing term distances) and a familiar `projection view' (a t-SNE projection commonly used by model developers). Interactions include adding model instances, filtering models based on multiple criteria, reordering (to find trends and outliers in model instances) and browsing and search features to isolate problems and new labels. \lvt broadly allows the user to identify relationships and correlations between parameters and output metrics \textit{and} to inspect the difference between the embeddings produced by each model instance. 

We demonstrate the utility of \lvt through a case study that replicates the early word2vec work as well as a performing a realistic user study using a real-world task. Participants were able to more rapidly and effectively tune a model using \lvt over default tools. As critically, we found that they were also better able to better understand the relationship of different hyperparameters.

In this work we contribute a \lvt, a novel visualization system for tuning hyperparameters of word embedding models. We believe that this approach can be adapted to other machine learning tasks where tuning is necessary.  Additionally, we demonstrate the usefulness on \lvt on a realistic task. 

%% file: 02_related.tex
\section{Related Work}
Work related to our project broadly falls into two broad lines of research and development (with some significant recent overlaps). The first line emerges from the machine learning and database communities on (non-visual) model management and tuning (e.g., ~\cite{modeltracker}). The second line of work is from the visualization and HCI communities on data-specific (i.e., text) visualizations and interactive learning.

\subsection{Hyperparameter Spaces}

\begin{figure}
  \begin{center}
    \includegraphics[width=0.5\textwidth]{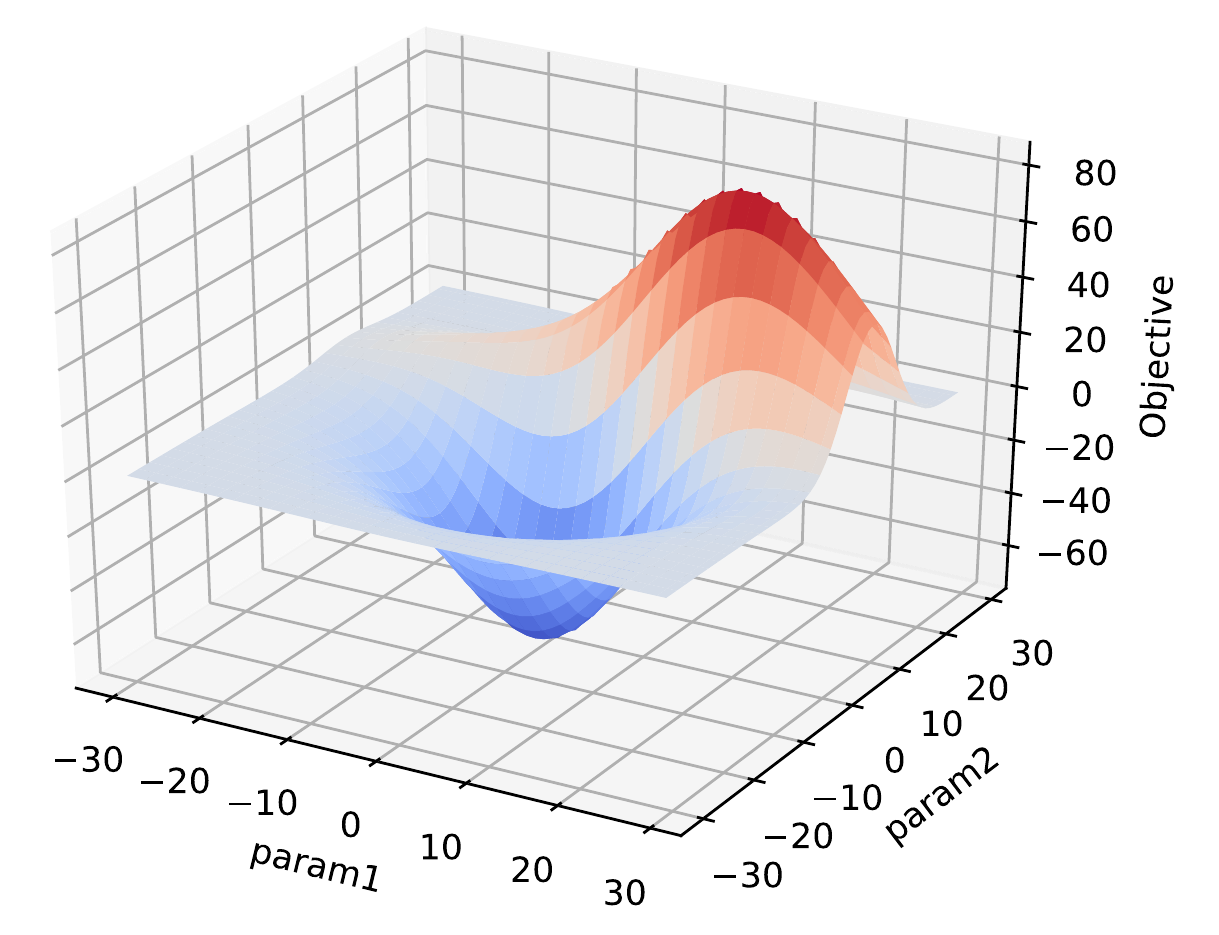}
  \end{center}
\caption[hyperparameter space]{3D plot of two hyperparameters.}
\label{fig:3d}
\end{figure}

However, some authors have adopted the use of static~\cite{Young2015} and interactive parallel coordinates~\cite{vizier} to illustrate the selection of hyperparameters. We take these as evidence that \lvt visualizations can be effective for explaining the relationship between hyperparameters. By augmenting generic parallel coordinates with domain-specific views, \lvt supports finer grain inspection of models. Many visualizations of hyperparameter spaces utilize static 3D models (see Figure~\ref{fig:3d}).

\subsection{Text and Machine Learning}
There is a large amount of work on use of visualization to inspect text, some with a machine learning focus. For example, AntConc provides concordances and other visual toolkits to support corpus linguistic analysis, such as word frequencies and collocation inspection~\cite{anthony2011antconc}. Kulesza et al. use simple bar graphs of feature importance to provide answers to the user's \textit{why}-questions regarding text message classification in an email client application~\cite{kulesza2011oriented}.  

Chuang et al. use association matrices and alignment charts to investigate latent topic coverage of a large collection of model variants~\cite{chuang2012termite,chuang2013topic}.
LDAvis employs interactive visualizations to facilitate the user to interpret the content and inter-relationships of different latent topics learned by a topic model~\cite{sievert2014ldavis}. Other examples related to topic visualization include \cite{cui2014hierarchical,liu2009interactive}. Alexander and Gleicher~\cite{alexander16} built a visualization system to compare different topic models. The tool visualized (in)stability of different topic-document connections but did not integrate views into the hyperparameter space and metrics to enable end-user driven optimizations.

These tools provide inspection capabilities for different text-related modeling tasks. They do not however, directly address the model tuning problem. In most cases, each visualization provides access to at most one model instance at a time.

\subsection{Visual Inspection of Neural Networks}

Visualization support for machine learning engineering, and in particular neural nets, has seen significant recent attention (surveyed in~\cite{hohman2018visual}). While model construction~\cite{8019861}, debugging and model selection have been identified as key tasks, hyperparameter tuning has seen less attention. While the attention on building and comparing new architectures is reasonable given the proliferation of new techniques, tuning of the type we describe here is critical to real-world deployment and adaptation of existing architectures to new contexts.

Model comparison is a critical step in model selection. However, many current approaches focus on comparison of the model output or internal `model states' rather than on the parameters that caused a certain behavior. Some exceptions include work on understanding the impact of input data on performance~\cite{Kulesza:2015:PED:2678025.2701399,Patel:2008:ISM:1357054.1357160}.

Using visualization techniques to improve understandability of neural network (NN) models has also drawn a great deal of attention. However, many NN visualization projects are focused on computer vision~\cite{yosinski2015understanding,samek2015evaluating}. There are also several interactive visualization projects developed for educational purposes, such as the recent Tensorflow Playground (\url{http://playground.tensorflow.org/}) and distill.pub~\cite{olah2018the}. More elaborate neural network architectures, such as convolutional neural networks, have also seen progress in using visualization to understand network behavior~\cite{DBLP:journals/corr/SimonyanVZ13, DBLP:journals/corr/ZeilerF13}. 

Compared to images, videos, or quantitative multidimensional datasets, words and sentences lack natural visual representation and can be harder to interpret. The required transformation of text to data that can be visually encoded presents unique challenges. To improve understandability of natural language representations learned by neural network models, existing techniques for visualizing high-dimensional data are often borrowed, such as principal component analysis (PCA)~\cite{jolliffe2002principal}, multi-dimensional scaling~\cite{kruskal1964multidimensional}, t-SNE~\cite{van2008visualizing}, and other multi-view analytics solutions~\cite{conceptvector}. While useful for inspecting particular embeddings, these encodings/systems do not address the problem of hyperparamter comparison and tuning. 

Notable work on visualizing specific text-based deep learning include systems by Karpathy et al. that employ multiple views of activation levels to illustrate how recurrent neural networks capture patterns in text sequences, such as long-range dependencies~\cite{karpathy2015visualizing}. Others have demonstrated the use of visualizations to investigate different NN features for both text and images (\cite{ming1understanding,bilal,datacube}), finding patterns in hidden states (\cite{DBLP:journals/corr/StrobeltGHPR16,deepeyes,liushi}) or analyzing classifier output (\cite{activis}). These approaches may be integrated into our own if \lvt is adapted to other neural architectures or ML models.

\subsection{Interactive Learning}
While most machine learning visualizations (including those focused on neural networks) target individual models, a few systems seek to provide inter-model comparisons. ModelTracker~\cite{modeltracker}, for example, allows the user to track the evolution of a model in terms of accuracy and provides interactive learning features. The tool allows the user to inspect  model-specific classification metrics and add new labels. This notion of interactive learning~\cite{amershi10, amershi2014power, LIU201748} inspired features in \lvt to add labels through the visualization.

%% file: 03_design.tex
\section{Task Analysis}

Based on our experience with early protoypes (\wevi, \lvo) and in producing educational material for word2vec users~\cite{rong2014word2vec} we identified key tasks and guidelines for hyperparameter tuning that motivate our design. We describe our prototypes and tasks below.

\subsection{Prototypes: \wevi and \lvo}
[Author note: we have anonymized the references and demo in our video but the `live' versions are not anonymous.]

\label{sec-lamvi:v1v2}

\begin{figure*}[!htbp]
\begin{center}
              \includegraphics[width=.8\textwidth]{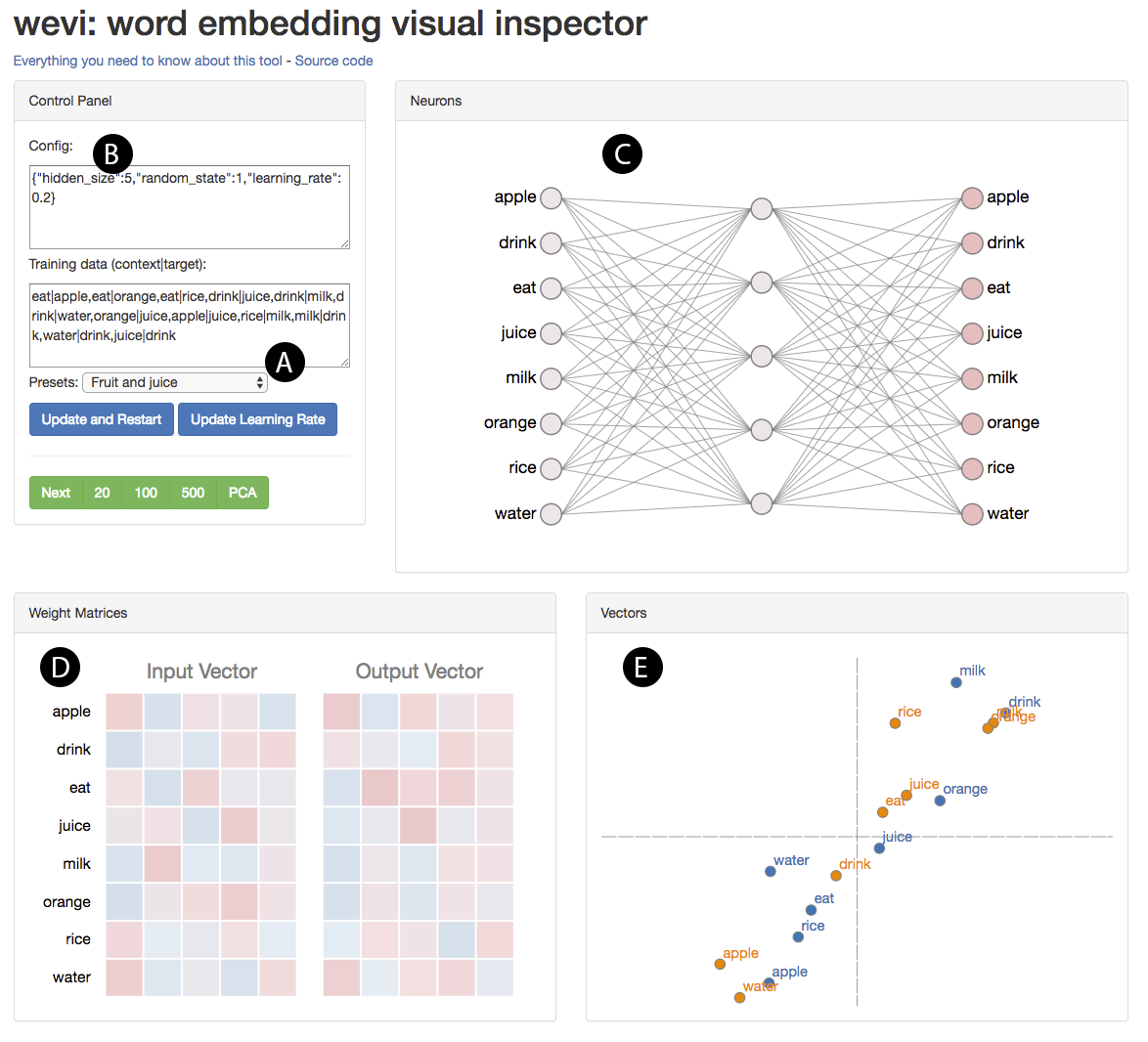}
\caption[Screenshot of \wevi]{The \wevi interface}
\label{fig:wevi-screenshot}
\end{center}
\end{figure*}

\begin{figure*}[!htbp]
\begin{center}
              \includegraphics[width=.8\textwidth]{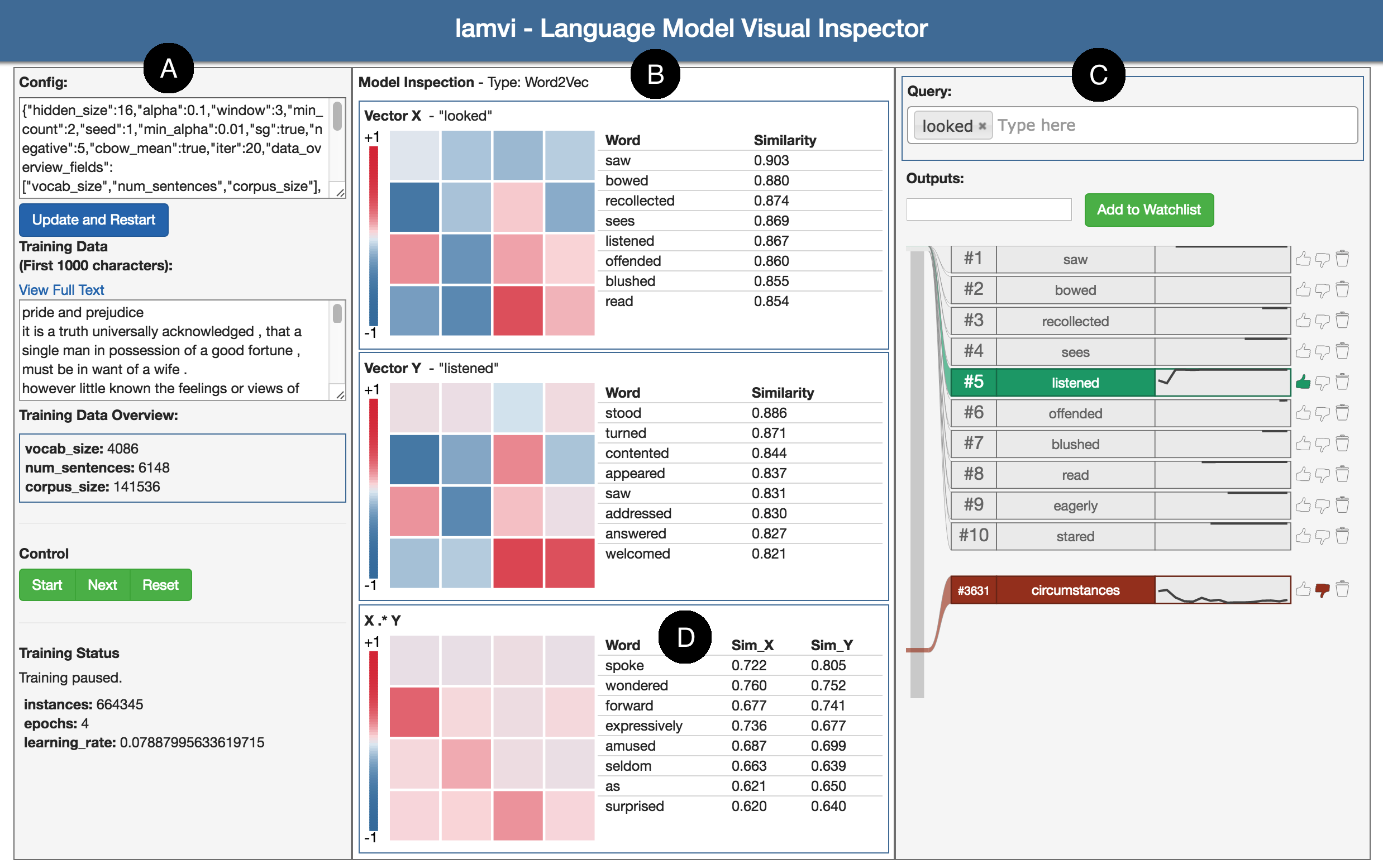}
\caption[Screenshot of \lvo]{The \lvo interface}
\label{fig:lamvi1-screenshot}
\end{center}
\end{figure*}
  
\lvt is the third in a line of visualization systems developed for analysis of word embeddings.  The first, \wevi~\cite{wevi}, was produced as a teaching tool to work alongside a (now-popular) tutorial on word2vec hyperparameter setting~\cite{rong2014word2vec} (see Figure~\ref{fig:wevi-screenshot}. While popular, \wevi was focused on extremely small language models and did not support model comparison. To test the impact of hyperparameters, the student would make a change and observe the effect on the neural network weights during training. Though useful for understanding how training happened and embeddings could be used to find close terms, the tool did not support realistic tasks. Nonetheless, we observed that end-users would iteratively vary the hyperparameters to understand their effect on the model.

While not intended for real-world use, \wevi did clarify for us that students, and likely developers, were not interested in vector weights for individual words. Rather, the relationship of words to each other was critical in determining whether a particular embedding was good. This focus on word \textit{relationships} has become a key guideline for our subsequent implementations.

\lvo~\cite{lamvidemo,rong2016visual} was built to support real-world developer practice. The \lvo interface (see Figure~\ref{fig:lamvi1-screenshot}) provided similar controls (panel A) to \wevi, though it used a full implementation of word2vec (allowing hyperparameters like window size to be controlled). Based on our experience with \wevi, we allowed \lvo users to focus their analysis to specific ``anchor'' terms (in the example the input text is Jane Austen's ``Pride and Prejudice'' and the anchor term is ``looked'') and the relationship of this anchor to other terms. \lvo allowed end-user to create compound vectors (e.g., \textit{elizabeth} $+$ \textit{bennet}) as well as displaying a `tracking list.' Tracked terms were those the end-user believed should (or explicitly should not) be related to the target term (a form of ground-truth labels). During training, \lvo would display the distance of the two terms (anchor and tracking) in a ranked list that simultaneously allowed the developer to identify convergence. Similarly to \wevi, the \lvo end-user could initiate training and observe the various visualizations dynamically change based on hyperparameter configurations.

\lvo solved a number of problems with \wevi. By explicitly tracking the interesting aspects of the model--the \textit{interactions} between terms, rather than the terms themselves--the tool supported both easier debugging and also allowed the end-user to dynamically build ground-truth labels based on their observations of the data. In using \lvo for our own projects and observing others' use, it became clear that one of the most common use-cases, comparing models, was not supported directly. Users of \lvo would make changes to the hyperparameters to retrain the model but would quickly lose track of the behavior of previous configurations. \lvt was specifically designed with this problem in mind.

\subsection{Design Guidelines}
Our experience with \wevi and \lvo, as well as in model building allowed us to identify three main decision points: \textbf{(D1) decisions on which models to test}; \textbf{(D2) decisions on how models are evaluated}; and \textbf{(D3) decisions on which models to use}. Each decision requires the execution of multiple analytical tasks.

\noindent
\textbf{D1: Model Construction}--Given the large hyperparameter space and cost of learning and testing, a key decision point in model construction is where resources are best expended. This decision is strongly connected to determining which model to \textit{use} (D3) but may involve a broader set of tasks.

\begin{itemize}[noitemsep]
    \item \textbf{T1}: Identify unexplored hyperparameter settings. To avoid overfitting, the developer should be able to identify which portions of the hyperparameter space are unexplored.
    \item \textbf{T2}: Identify the correlation and interaction between hyperparameters and output metrics. A key type of analytical determination is the identification and mental modeling of those interactions that display trends. Specifically, the developer must know if tuning the hyperparameter value(s) \textit{in a specific direction} will produce better results.  
    \item \textbf{T3}: Identify satisficing hyperparameter setting ranges. As hyperparameters often lead to tradeoffs, a developer should be able to identify those models that satisfice their need (given evaluation metrics) and focus on those for construction and selection.
\end{itemize}

\noindent
\textbf{D2: Model Evaluation}--Model are principally selected, or trained, based on different metrics (i.e., pre-established benchmarks). These can include temporal performance measures (e.g., speed of training or execution), accuracy metrics (based on model-specific ground truth or an evaluation of the model's contribution to downstream applications), and other model features (e.g., memory footprint, power consumption, etc.). It is our goal for developers to be able to identify satisficing ranges for multiple performance metrics (T3). Additionally, because ground truth test-sets are often specific to the training data or generic (e.g.,~\cite{semeval}), they often require adaptation. When applying the model to new training data, the developer often needs to change the ground-truth. Put another way, the developer should not only be able to define which performance metrics to apply (with their given ground-truth labels), but also dynamically augment ground-truth labels.

\begin{itemize}[noitemsep]
    \item \textbf{T4}: Identify, and mentally-model tradeoffs in hyperparameters given different performance metrics.
    \item \textbf{T5}: Identify satisficing-models given global performance metrics.
    \item \textbf{T6}: Compare low-level model differences (i.e., word or word-cluster specific) to identify new examples to incorporate into performance metrics.
\end{itemize}

While we focus in this work on text embeddings, it is worth acknowledging that many of our ground-truth labels are `second order.' That is, there is not `correct' vector representation for any particular word. Rather, the user might want a pair of words to be similar ('close'), or some order on nearest neighbors, or may want a certain group of words to form a coherent cluster. Conversely, the user might have negative examples and want to exclude certain words from a cluster. The forms of ground-truth can vary significantly depending on task. The visualization interface must make it easy for the user to explore the possible candidates for establishing such ground-truth items (T6), and to manage them by creating, updating, and deleting individual items.

\noindent
\textbf{D3: Model Selection}--A final step in our process is to identify the model to deploy. This involves a number of the tasks described above (T2, T3, T5, and T6). In the act of isolating `good' models, and having understood their trade-offs and limitations, a decision can be made about the `best' model. This step also allows the developer to document their choice in a way that others can understand the decision.

\begin{itemize}[noitemsep]
    \item \textbf{T7}: Document model choice/selection for external audiences.
\end{itemize}

%% file: 04_system.tex
\section{LAMVI-2}
\label{sec-lamvi:system}

To support the tasks described above, \lvt is composed of a number of interlinked views and interaction patterns. We briefly decompose the main elements (Figure~\ref{fig:lamvi2-screenshot}) in the context of the high-level decision, and low-level analytical tasks.

\subsection{Parallel Coordinates (panel A)}
Understanding hyperparameters and metrics is inherently a multivariate data analysis problem. There are quantitative, ordinal, and categorical hyperparameters, and similarly quantitaitve, ordinal and categorical metrics. The analytical tasks that involve comparing (T1-T5) inherently require multiple comparisons between these variables. Additionally, many key decisions in our workflow require filtering on multiple dimensions simultaneously (T6). The interactivity that is afforded by parallel coordinates fits well for the goal of identifying satisficing models. 

Parallel coordinate plots~\cite{inselberg1985plane} naturally fit as good mechanism for encoding multiple dimensions onto a 2-D plot. There are multiple reasons to adopt this form beyond the expressiveness of the encoding given our task. First, parallel coordinate plots are becoming familiar to end-users. Despite the learning curve, they are effective representations and are already being adapted in the context of model comparison. For example, Google's Vizier~\cite{vizier} and the open-source Ray Tune software~\cite{raytune} offer parallel coordinate plots. This use is somewhat unsurprising as one of the original case studies~\cite{detective} describes the use of parallel coordinates to trade-off parameters (of chip production system) against different performance measures. While there are likely better static representations, interactive documentation (T7) can be achieved by providing a pre-populated version of \lvt. 

This view is attractive for our context as the encoding supports rapid identification of positive/negative relations (T2, T4) between data dimensions (based on slope). While a limitation of the technique is that, at any given time, each dimension can have at least two neighboring dimensions, but most parallel coordinate implementations (including our own) allow coordinates to be re-ordered. Since one of our requirements above is to facilitate efficient identification of relationships between model hyperparameters and performance scores, we opt to use parallel coordinate plots to depict both kinds of variables. \textit{To begin,} hyperparameter dimensions are placed on the left and output/performance dimensions are placed on the right (with the heatmap in the center). This reflects the natural ordering of parameter to model to outcome but elements may be reordered if necessary. Unexplored hyperparameter spaces (T1) are salient as `empty' areas in the parameter axes.

Our implementation supports the standard parallel coordinates interaction patterns.  By brushing along a specific axis, the end user can \textbf{filter} the data to narrow down the number of lines being highlighted to a smaller set whose crossing points with the axis are within the brushed range. To focus, the brushed region can be rescaled or moved through cursor interactions. Double clicking an an axis label with cause the data to \textbf{flip} for that dimension. This feature may help the user reduce the number of line crossings between axes and also allows for consistent mental models. Other features such as colors and bundling can be enabled, but are not done so by default. Together, these features enable the filtering necessary for identifying good models (T3, T5).

While \lvt can support any number of loaded models, realistically we do not expect more than 30-40 (and that depends significantly on model training costs). In part, an appeal of \lvt is that we can `seed' the system with random points in the hyperparameter space and allow the end-user to narrow this space by visual inspection.

\subsection{Embedding Explorer (panel C)}
Embedding words in a 2D space is a natural way to inspect models. In particular, for task T6, inspecting the words in this way allows the developer to identify possible issues with the model. More specifically, approaches such as t-SNE~\cite{van2008visualizing,wattenberg2016how} and PCA reduce multi-dimensional vectors to two dimensions, allow the developer to inspect the relationship of terms both in magnitude and distance. For better or worse, such representations have become standard for both interactive and static visualizations (e.g., Google's Tensorboard: https://www.tensorflow.org/programmers\_guide/embedding). The benefit of projections is that the viewer is that related terms are placed near each other in the low-dimension projections. Gestalt properties of this projection make it easy to find patterns and detect outliers. For neural-net based embeddings, a further (argued) benefit is that the  \textit{relative directions} of terms has `meaning.' 

Traditional use of 2D projections involves the placement of all the words in the model or some large top-$k$. Unfortunately, with many words, the visualizations can become difficult to navigate. More critically, there are more opportunities with large differences for words to be placed near each other as a consequence of the dimensionality-reduction and not due to `true' distance~\cite{chuang12}. Projecting all the terms has the appeal that it follows the `overview first, filter, details on demand' mantra. However, it is unclear to us that the particular `overview' provided benefit for the tasks we outlined. 

While \lvt can project all terms in the model, we employ a different default strategy. Instead of a view of the entire set of words, the \lvt Embedding Explorer shows a projection anchored on a specific query 'term' and a set of its nearest neighbors. As different models in the parallel coordinate view are selected or loaded, the projection updates. \lvt uses t-SNE for projection. The interactive explorer supports basic pan-and-zoom functionality to explore the projection. The 'focus' term is made salient by using a larger font size (and is orange). Terms that are labeled (e.g., antonyms or synonyms) are also made salient by color coding and increased font size. The developer can easily find term pairs that are unlabeled as well as the clustering of labeled examples (e.g., are the antonyms all near each other?). In the event that these labeled terms are not part of the k-nearest neighbors (i.e., an antonym \textit{could} be very far), \lvt will add those terms to the projection.

In \lvt, the developer can perform tasks such as filtering, sorting, creating new models, or switching to old ones. Terms in the embedded view are animated so the developer can observe the impact of the interaction on the spatial relationship of the terms. A secondary benefit of focusing on a small set of terms is that these animations can be visually tracked more easily. Terms that are no longer `near' to the query are animated out, while those that now meet the criteria are animated in. Label placement also dynamically changes based on the vector representations of the selected model. The t-SNE optimization algorithm randomly initializes the embedding vectors and often the labels temporarily shrink to the center of the space, making it difficult to track terms as they move. To counter this, we first perform 150 iterations of the t-SNE algorithm in the background to find a better starting state for the animation.

As we describe below, the t-SNE projection is in some sense a `dual-encoding' of term distances (though it uniquely displays direction). We have found that a useful interaction pattern is to support a focus on word pairs or sets (the smallest 'unit' of useful analysis for word embedding). By clicking to highlight two or more words, and switch focus between models, the developer can track the relationship of a small set of terms between model variants. Because highlights are preserved between views, the user can both visually track where the words are relocated and acquire additional spatial awareness of changes in the embedding space across different models. This allows them to identify patterns that are consistent (building a mental-model of the models' behavior) as well as finding models that are `outliers.'
. 

\subsection{Nearest-Neighbor Heatmap (panel B)}
A unique feature of \lvt is the Nearest-Neighbor Heatmap (panel B). The heatmap is connected both to the parallel coordinates plot (models are rows) and the t-SNE projection (words are columns). Words are both visually linked (through a line) as well as through linked brushing. Mousing over a word will highlight the appropriate column.

The heatmap is specifically targeted at task T6, comparison of low-level model differences. The encoding allows us to compactly display data with a high information-density summary~\cite{munzner2014visualization}. Each cell corresponds to the distance between a `query' and a sampling of related terms.  For example, in Figure~\ref{fig:lamvi2-screenshot}, the current query term is ``legitimate'' (also displayed as the first column).  The second column corresponds to the term ``dubious.''   

As the number of nearest neighbors (i.e., columns) and models increases it becomes difficult to identify patterns and outliers in the data.  We use a combination of visual encodings and interaction techniques to help the user navigate this information. Patterns of particular interest are those columns where models agree on the distance between the query term and the other word.  For example, the models seem to agree that `valid' is similar to 'legitimate.'  This manifests as a column of a single color. A second pattern of interest are those when there is either a single outlier in a column or a bifurcation, where some models place the term 'close' (blue) and some 'far' (red). For example, five of the seven models place 'moderate' (4th column) far and two place it close. Other patterns delineate outliers and clusters at the model level. By looking at multiple rows, the end-user can find `clusters' of models that generate similar embeddings and those that are distinct or anomalous.

To increase the number of columns that can be simultaneously display we designed a two-mode display for columns inspired by Tablelens~\cite{rao1994table}. This increased the amount of data displayed but still supported the pattern finding task. Under the ``normal'' mode, columns in the matrix are compressed to a few pixels (for example the column for 'valid'), minimizing space but still allowing the viewer to identify patterns across all models. Under the ``zoomed-in'' mode, the columns in the matrix are displayed with a larger width with labels displayed below the bottom row. By default, the top-K nearest neighbors of the query word will have their corresponding columns zoomed-in (K starts at 15) . As the user interacts with the other parts of the interface, pertinent columns will be adjusted to zoom-in. This focus switching mechanism optimizes the efficiency of space usage while allowing the user to quickly navigate through the most relevant columns. Other heat-map `modes' can put columns with inter-model agreement in the 'normal' (low-pixel) and low-agreement terms in the zoomed-in state.

Although the heatmap is integrated into the parallel coordinates display, the vertical ordering of the rows does not necessarily correspond to the same high-low ordering. While this breaks the encoding, it also allows us to dynamically reorder both the rows and columns of the heatmap to emphasize patterns. Line segments connect the heatmap rows to the parallel dimensions coordinate on the left and right ensure that the viewer can `follow' a particular model's encoding.  When models are filtered in the parallel coordinate dimensions, the corresponding rows are highlighted (those retained by the filter) or are `grayed out.' In our experience, we have not found the `interruption' of the heatmap in the middle of the parallel coordinates plot to be disruptive (visually or semantically). However, we note again that  parallel coordinates axes that are classically 'dimensions' of the encoding can be grouped together (i.e., hyperparameters and metrics), placing the heatmap to the side. This may be a desirable configuration for some end-users or for specific model types.

\begin{figure*}[t]
\begin{center}
\centerline{\includegraphics[width=.8\textwidth]{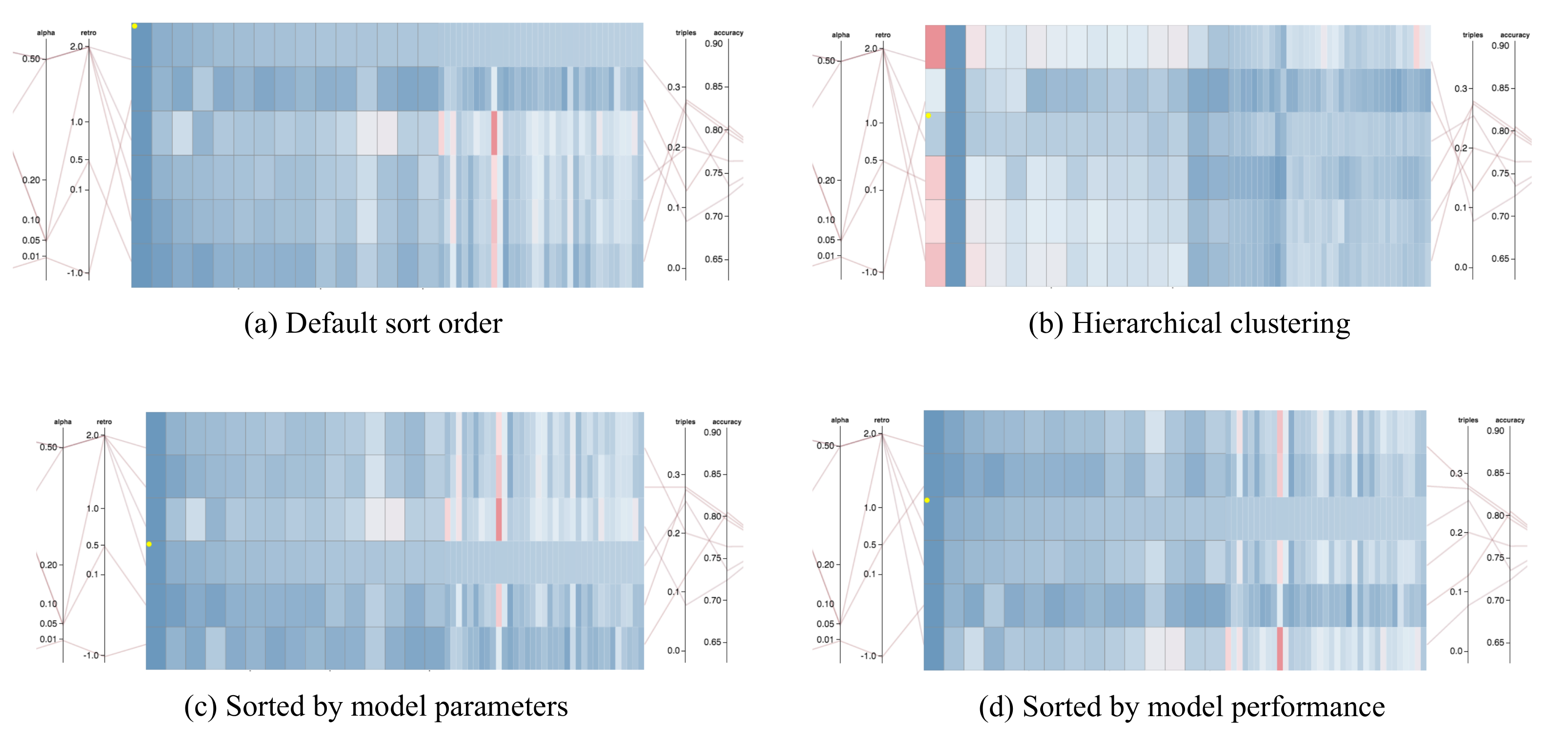}}
\caption[Four sorting modes of \lvt heatmap.]{Four sorting modes of the \lvt heatmap: (a) default mode -- models are sorted by loading order; (b) cluster-sort mode -- models and words are sorted according to the dendrogram of hierarchical clustering; (c) hyperparameter-sort mode -- models are sorted by the right-most hyperparameter in the parallel coordinate plot; (d) metric-sort mode -- models are sorted by the left-most performance metric.}
\label{fig:lamvi2-four-matrix}
\end{center}
\vskip -0.2in
\end{figure*}

\lvt supports a variety of \textbf{clusters} and \textbf{orderings}. As the primary focus for our heatmap is for the identification of model disagreement and clusters, the clustering of models can simplify the process greatly. Figure~\ref{fig:lamvi2-four-matrix} illustrates four different configurations for the same data.  The default sort order places rows in the order the models were loaded.  To minimize edge-crossings, the heatmap can also be sorted by model hyperparameters (Figure~\ref{fig:lamvi2-four-matrix}c and d). In this case, \lvt takes into account the value of the hyperparameter dimension immediately before (in this case 'retro') or the performance metric immediately after ('triples') in deciding the order of the rows. Because the coordinates can be reordered, the user can move a metric dimension next to the heatmap and order by this value. This will cause the rows of the heatmap to reorder to correspond with the metric and allow the user to inspect the properties of 'good' versus 'bad' models.

A more sophisticated order is created through hierarchical clustering. We achieve this by running a hierarchical clustering algorithm on the heatmap data (row-wise). The order for the rows corresponds to the order used to visualize the dendrogram. The effect is that models that have similar distance properties are placed near each other. As illustrated in Figure~\ref{fig:lamvi2-four-matrix}b, patterns become apparent (e.g., the agreement in the last three rows) or that the top row is an outlier. Clustering in this way has the further advantage of making salient specific cells that are different (i.e., within-cluster outliers).

\begin{figure*}[t]
\begin{center}
\centerline{\includegraphics[width=.8\textwidth]{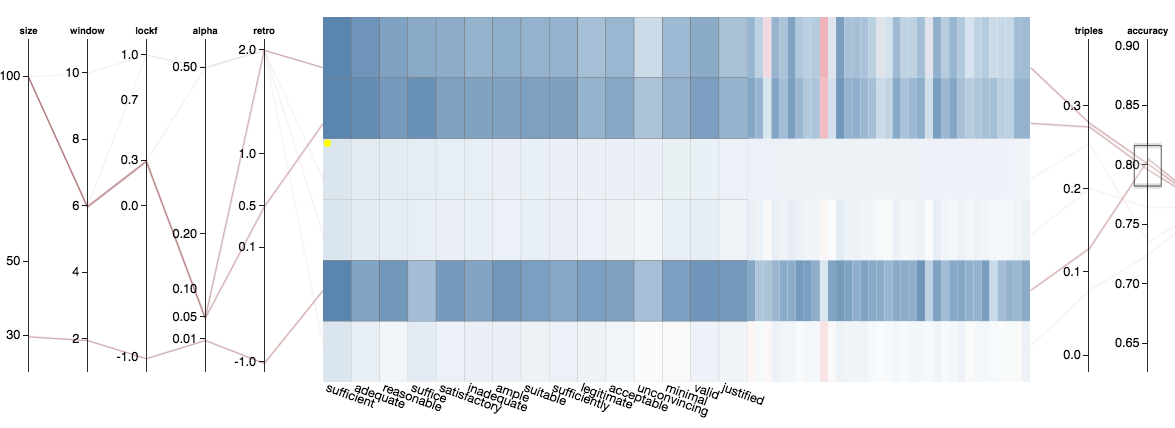}}
\caption[Filtering of rows in \lvt heatmap via parallel coordinate brushing.]{Heatmap rows can be filtered and highlighted by brushing axes on the parallel coordinate plot. This allows a focused comparison among models that meet a user-specified filtering criterion.}
\label{fig:heatmap-filter}
\end{center}
\end{figure*}

We retain the \textbf{filtering} interaction pattern of the parallel coordinates inside of the heatmap (see Figure~\ref{fig:heatmap-filter}). Whenever the user applies filtering or brushing to one of the dimensions in the parallel coordinate plot, the corresponding rows in the heatmap will be highlighted. The non-highlighted rows are still visible but will be applied with a semi-transparent mask. Conversely, the end-user can select rows in the heatmap to filter one or more models. 

\subsection{SPLOM}
A fourth panel (not pictured) displays a Scatter Plot Matrix (SPLOM). The grid displays pairwise correlations between all hyperparameters and metrics (each point is a model instance). Because it is sometimes difficult to quantify correlations between dimensions in parallel coordinates directly, the SPLOM view provides a more accurate and comprehensive encoding of this data (further supporting T1-T4).  The SPLOM is interactive and linked to the other displays. As the end-user filters (by drawing selection rectangles), models are dynamically filtered from all the views.

\subsection{Inspecting Models and Training}
Recall that a key goal for \lvt is to facilitate the creation of training data (T6). The end-user may issue queries through a text-input dialog. This will refresh the embedding view to include the query string and nearby terms. When the user is interested in the similarity and neighborhoods of a \textit{pair} of words simultaneously, she may directly query the two words together. This allows her to more easily mark good cases and bad cases as more words are displayed (neighbors of both terms) that likely match the relationship of the query pair.

Our current implementation also supports inspecting the emergence of linguistic regularity captured by the model. The user may enter queries like ``king --queen woman'' and observe how the desired candidate, ``man'', evolves. The user may also inspect the activation levels of the hidden units given all three words as context. The distances between query and neighbors is shown automatically in the heatmap.

\begin{figure*}[t]
\vskip 0.2in
\begin{center}
\centerline{\includegraphics[width=.8\textwidth]{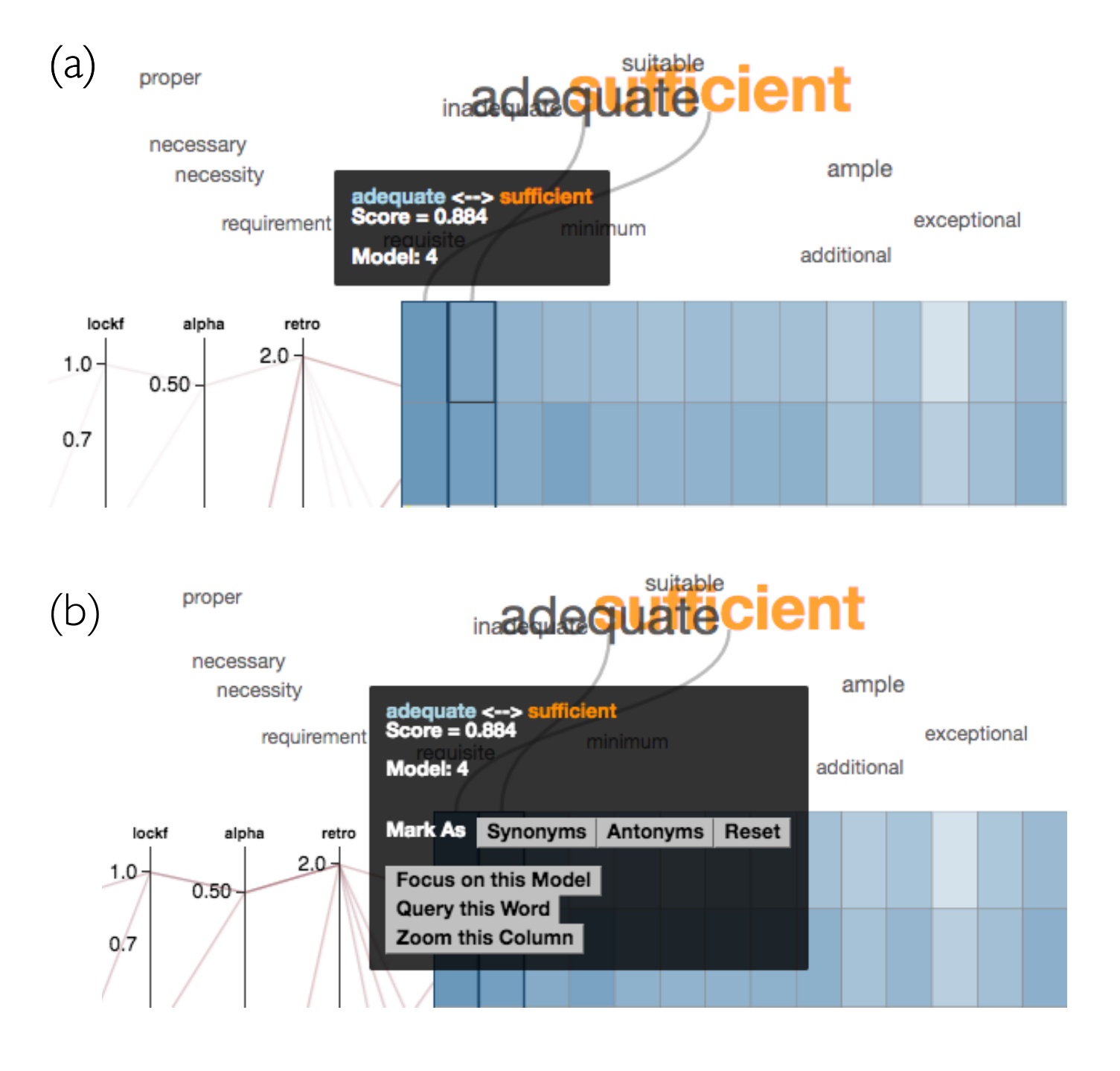}}
\caption[Creating user-customized ground-truth in LAMVI-2.]{Multifunctional tooltip in LAMVI-2 heatmap: (a) compact tooltip shows the similarity of the word pair under focus; (b) expanded tooltip allows the user to assign a label to the word pair, creating user-customized ground truth.}
\label{fig:lamvi2-tooltip}
\end{center}
\vskip -0.2in
\end{figure*}

Clicking on a cell in the heatmap will show a dialog box by which pairs can be `labeled' (see Figure~\ref{fig:lamvi2-tooltip}). Though we can support multiple labels, the default prototype categories include \textit{synonyms} and \textit{antonyms}. As labels are created, these are dynamically fed as labels to downstream evaluation `modules.'  

A particular workflow that can be executed is the identification of models that correlate with outliers. When the user explores the embedding and discovers words that are inappropriately placed in the neighborhood of a query she can click on the word label to automatically zoom into the associated column in the heatmap. The user can scan down the entire column and see if the score is consistently spuriously high. She can then filter on those rows to see if certain hyperparameters result in the undesired score.

\subsection{Implementation Details}
The back-end of the \lvt system is written in Python and was designed with extensability in mind. The front-end was programmed using D3~\cite{bostock2011d3}. A model developer can configure and evaluate models programmatically. These models can then be loaded into the interface. We currently provide basic command line tools that will accept a high-level configuration (e.g., one indicating the space of hyperparameters to explore) and builds a model for each point in this space. A web-based tool allows for generation of a JSON configuration file that can be fed to the command-line pipeline.  The tool is currently configured for multi-scale grid search as well as random search~\cite{bergstra2012random}, but can be readily adapted to other hyperparameter optimizers~\cite{NIPS2011_4443,NIPS2012_4522}. Labeled examples created by the developer are stored in a text-based label file. The entire `state' of a particular experimental run can be preserved and provided to other developers as documentation (T7).

%% file: 05_casestudy.tex
\section{Case Study}
\label{sec-lamvi:casestudy}

To understand how \lvt could be used in a realistic situation we replicated the work by Mikolov et al.~\cite{mikolov2013efficient} (as implemented by Rehurek for the GenSim package using the \textit{text8} corpus~\cite{rehurek}) using our tools. In part, we selected this case as both the the original Mikolov et al. paper and the implementation by Rehurek report their parameters without specific discussion of trade-offs and optimality. We demonstrate both that it is possible to identify the `optimal' hyperparameters but also that we can more fully describe the effect of these parameters on performance. An advantage of using \lvt is that it becomes easier for the author to identify and report (task T7) on patterns in the data that may provide guidance for others using new neural architectures. As the architectures, and associated hyperparameter spaces, become more complex, such guidance is critical.

To build a set of comparative models we used hyperparameter ranges that were expansive, but included the `optimal' values reported in the studies above. In the interest of replication, merging with other models and retrofitting were not used during training. Hyper-parameter ranges were: Word Vector Size: 100--400, increments of 50; Skip-Gram architecture: Both with/without; Hierarchical Softmax: Both with/without; Window size: 3--7, increments of 2; Negative samples: 5--20, increments of 5. 

\begin{figure*}[t]
\begin{center}
\centerline{\includegraphics[width=\textwidth]{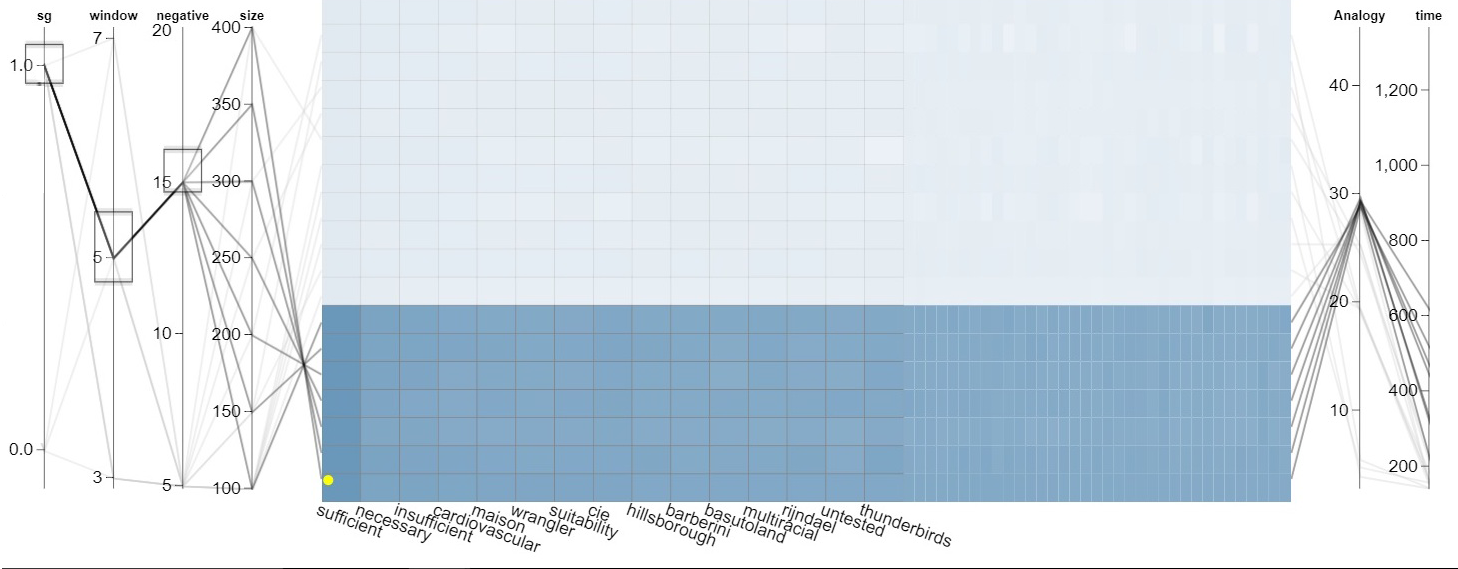}}
\caption[Inspecting size hyper-parameter in \lvt]{A view of the \lvt interface for the word2vec case study. The end-user has targeted the word 'sufficient' and has filtered to those models containing skip grams, a window of 5, and 15 negative samples.}
\label{fig:LAMVI-size}
\end{center}
\end{figure*}

Figure~\ref{fig:LAMVI-size} displays the \lvt interface with the different models in this space active. The analogy metric is displayed on the right as is training time. When filtering on the maximal accuracy we achieve results comparable to Rehurek, including roughly 28\% accuracy on the analogy task for the model with settings \{size: 200, skip-gram, negative sampling w/ 15 samples, window size: 5\}. More critically, we observe trade-offs that were not described in the original paper, but now understood in the outside literature. For example, the use of both skip-gram and hierarchical sampling results in increased accuracy at the cost of a significant increase in training time. 

We find that these optimal parameters are slightly different than those reported by Mikolov (dimension: 300, window size of 7). This may be partially due to the use of \textit{text8} but demonstrates the need to perform algorithmic tuning when changing corpora. Additionally, given the time to train different sets of models, performance improvements can be achieved through more fine-grained tuning of hyperparameters. However, we do find that the granularity of the adjustments correlates with the granularity of the improvement. The SPLOM visualization further shows decreasing returns for more training time relative to analogy performance.

%% file: 06_eval.tex
\section{Evaluation}
\label{sec-lamvi:experiment}

Evaluating tools for deep learning models can be difficult, as they require significant participant expertise. Realistic scenarios are similarly complex, as a local evaluation of the embedding may not be sufficient. For example, one might create a classifier for synonyms and use this in a downstream information extraction (IE) system. An optimization in the sub-module (the synonym classifier) may not mean optimal behavior in the IE system. To simulate this, we created a two-stage task for polarity classification of IMDB reviews (i.e., are reviews positive ore negative). Because we may not know the polarity of every word in the document, one approach is to find synonyms to known positive terms and label these as positive and identify antonyms and label them as negative (conversely, synonyms of negative polarity terms can be labeled negative).  This scenario naturally induces two metrics: the accuracy in identifying synonyms/antonyms and the accuracy in classifying reviews.  Our experiment focuses on how well (and quickly) participants can achieve an optimal hyperparameter-configuration using \lvt versus baseline ``command line'' style tools.

\subsection{Task and Model}
For our task, we used the Cornell IMDB movie review corpus~\cite{pang2002thumbs} as the training corpus (100k reviews, 50k of which have positive/negative sentiment labels). The input feature to this classifier is the average word embedding (i.e., the centroid of all words) in that document.  The intuition is that if the embedding is `good', positive synonyms will be grouped together and negative synonyms will be grouped together in space. If the document has more positive terms, the average vector will be in a `positive' location.  We train a logistic regression classifier for this ``downstream'' application.   We calculate $f_A$, as the accuracy score for the sentiment classifier using 50k documents (equal partition, positive/negative). The participant's goal is to provide the classifier with a good embedding. 

As described above, we may only have sentiment scores for \textit{some} of the words in the corpus.  By creating an embedding that places synonyms of a given polarity together, it becomes possible to provide scores for all words.  We challenge our participants to tune a word2vec skip-gram model~\cite{mikolov2013distributed} to provide such an embedding. We provide the participants with `synonym-antonym triples'--three words where one word is an `anchor' (e.g., \textit{ghastly}), one is a synonym (\textit{awful}), and one is an antonym (e.g., \textit{delightful}).  We also provide a few examples (specifically five) to the study participant (we call this the 'training set') with the expectation that they add additional examples.  Thirty-five example triples are withheld from the participant (the 'test set') and we evaluate whatever embedding configuration they pick against this set. The 40 triples (5 train, 35 test) were generated by consulting external lexicons (SentiWordNet~\cite{baccianella2010sentiwordnet}, \textit{Oxford American Writer's Thesaurus}~\cite{lindberg2012oxford}, and an online thesaurus dictionary.\footnote{\url{http://www.thesaurus.com/}.}). We specifically selected words with obvious polarity (top 10\% of SentiWordNet) so it was clear from our examples whether the sense was positive or negative. Only words that occurred between 150 and 250 in the IMDB dataset were used (to ensure sufficient data).

The \textit{triples} score, $f_T$, measures how well a model distinguishes synonyms from antonyms. It is calculated as the average difference between the similarity difference of a set of synonym-antonym triples:
\begin{equation}
f_T = \frac{1}{N}\sum_i^N(\cos(q_{i,A},q_{i,B})-\cos(q_{i,A},q_{i,C}))
\end{equation}
where $q_{i,A}$ and $q_{i,B}$ are the vectors of a pair of synonyms and $q_{i,A}$ and $q_{i,C}$ are the vectors of a pair of antonyms.

We use the average of two metric scores, triples and accuracy, to measure the performance of the model. For both $f_T$ and $f_A$, the user can access the corresponding scores assessed on the training set in the interface (i.e., for each parameter configuration a model was generated, and for each model accuracy was calculated for both the triple task and sentiment analysis task and provided to the participant). The final evaluation of the model was to be assessed on the test set.

\subsubsection{Parameter Space}
To control the scope of the experiment, we set some of the parameters to be non-adjustable by the user (specifically, we fixed the frequency downsampling parameter to be $1\times10^{-4}$, 5 iterations, and 5 negative samples). We allowed the user to select whether to have the vectors pretrained on an external corpus (the `lockf' parameter and whether to use an external lexicon to retrofit the vectors (the 'retro' parameter). Pretraining provides a `starting' point for training, for example based on Wikipedia, and is used when the domain corpus is too small. Retrofitting happens after the model training and is shown to be able to improve the quality of the vectors~\cite{faruqui2014retrofitting}. In our case, the paraphrase database (PPDB)~\cite{ganitkevitch2013ppdb} is used as the semantic lexicon for retrofitting. The database contains 8 million English lexical paraphrases (e.g., \emph{jailed} and \emph{imprisoned}). 

The parameters that the participant could vary included word vector \textbf{size}; sliding \textbf{window} length; whether vectors pretrained on Wikipedia should be used (\textbf{lockf}, which could be varied between 0 and 1 to determine the weight of the pretrained vectors on the final vectors, or -1 to be disabled); the initial learning rate for gradient descent (\textbf{alpha}); \textbf{retro}fitting, indicating whether to use an external dictionary to augment the vectors after training (ranging between 0 and 2, where a larger number indicates less external influence and a -1 disabling this step). While these parameters may appear overwhelming, we emphasize that they are realistic decision points encountered by developers.

\subsection{Procedure}
Participants answered a brief questionnaire about their experience and familiarity with machine learning and word embedding models. Each participant was randomly placed into a baseline condition or one in which they used \lvt. 

The \textit{baseline} emulated the command line tools currently available, and allowed the participant to enter model parameters and see the evalution metrics. The baseline also allowed the participant to enter `queries' into their model to see similar terms. Baseline participants could add triples to their own `test' set. While this is a `weak' baseline in that we did not offer any visualization tools, the command-line tools were familiar to many of our participants and were part of the `standard workflow.' Participants in the \lvt condition were given access to the full interface (with the exception of the SPLOM view that had limited use given the constraints below).

Models covering a broad parameter range (capturing all likely selections by the participant) were pre-generated.  Given the time constraint of the experiment, this would allow a participant to load a range of models without waiting for training.  Two models were preloaded into the experimental interface. Participants could load at most 13 more models at a time (for 15 total). This was done to create some `cost' to loading additional models as otherwise the participant could simply load all models and select the optimal one. In total, 374 models were pre-trained for the experiment.

After a three minute introduction using a prepared script, the participant was allowed five minutes to experiment with the interface and ask questions. Logging of interactions began and recorded which models were loaded, filtering operations, search, etc. The participant selected one model as the final submission. Finally, Each participant answered a questionnaire with questions taken from the system usability scale (SUS) as well as a few open-ended interview questions.

\subsection{Results}
Twelve subjects were recruited and received \$15 for participating. All were graduate students in computer science or a relevant major at a large public university with some machine learning experience. From our pre-survey, 25\% of the subjects reported to be beginner level with regard to machine learning, and 75\% self-identified as intermediate to expert level. One third of the subjects had learned about word embeddings but had not used them in practice; one third used them as a black-box component in a project; and one third had done at least a research project on word embedding models.

Table~\ref{table:lamvi2-eval} summarizes the results for the comparison between \lvt and the baseline system. On both the training set and the test, the average performance of the models in groups with the baseline and the \lvt system are reported. \lvt users consistently outperformed the baseline system under the metrics of both $f_T$ and $f_A$.

\begin{table}[htbp]
\centering
\caption[LAMVI-2 Evaluation Results]{LAMVI-2 Evaluation Results}
\label{table:lamvi2-eval}
\begin{tabular}{c|c|c|c|}
\cline{2-4}
\multirow{2}{*}{} & \multicolumn{3}{c|}{Training Set} \\ \cline{2-4} 
 & triples $f_T$ & accuracy $f_A$ & average \\ \hline
\multicolumn{1}{|c|}{Random} & 0.202 & 0.751 & 0.477  \\ \hline
\multicolumn{1}{|c|}{Baseline} & 0.277 & 0.795 & 0.536 \\ \hline
\multicolumn{1}{|c|}{\lvt} & \textbf{0.295} & \textbf{0.814} & \textbf{0.554}\\ \hline
\multirow{2}{*}{} & \multicolumn{3}{c|}{Test Set} \\ \cline{2-4} \hline
\multicolumn{1}{|c|}{Random} &  0.159 & 0.745 & 0.452 \\ \hline
\multicolumn{1}{|c|}{Baseline} &  0.203 & 0.791 & 0.497 \\ \hline
\multicolumn{1}{|c|}{\lvt} &  \textbf{0.223} & \textbf{0.810} & \textbf{0.517} \\ \hline

\end{tabular}
\end{table}

\begin{figure*}[htbp]
\begin{center}
\centerline{\includegraphics[width=\textwidth]{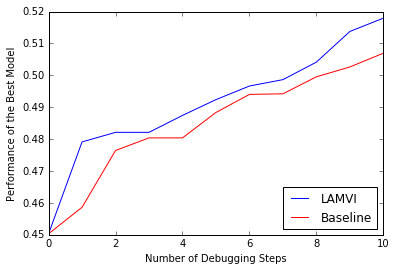}}
\caption[Model performance improvement process.]{Model performance improvement against the number of models loaded.}
\label{fig:curbest-models}
\end{center}
\end{figure*}

Figure~\ref{fig:curbest-models} compares \lvt with the baseline system in terms of the performance score ($\frac{f_T+f_A}{2}$) for the current best performing model at each interaction step (i.e., how many models were tested). We note that \lvt participants were able to improve model performance at a faster rate. 

Participants using \lvt generated 16 synonyms and 13 antonyms (on average) using the query functionality. We compared the performance of the user's best-selected model on each of their own synonym/antonym selections (simulating ground-truth construction) with the performance of their model choice on our hold-out set. The Pearson correlation across all participants (averaged) .599 and ranged from .509 to .774 (all statistically significant at $p<0.0001$). This indicates that participants `flagged' synonym-antonym triples that exhibit similar similarity/dissimilarity patterns as our preselected triples in the held-out dataset. Additionally, a higher correlation ($r$) generally means that the final user-selected model has a better task performance overall.

\subsection{Additional Observations}

\textit{Navigation Strategies}--Participants in the \lvt condition developed specific strategies to guide themselves to navigate through the parameter space. They noted that the parallel coordinates plot helped with this search, especially for the parameters that they were unfamiliar with.  For example, during the post-study interview, user P3 mentioned that ``I have a search algorithm. I start with a default setting, and vary each parameter at a time, and see if I can get the biggest improvement. Then I fix the one with the biggest improvement, and vary other parameters one by one.'' User P7 said, ``I first applied the prior knowledge that I have. Then I try the medium values for the options that I don't know, and then test the extreme values deviated from the medium values.'' Similar to these two users, most other users developed a variant of ``greedy search'' or ``gradient descent'' algorithm for parameter searching. 

For the \lvt participants the parallel coordinate plot was important to these strategies. P5 commented, ``It was tremendously helpful to use the parallel coordinates to tune down the scope of searching. I first look right to the pick the performance I want, and look left to confirm the intuition. I confirmed my intuition that lockf and alpha and retro should be locked to a specific value, and window and size can be fine-tuned.''

\textit{Outlier Model Detection}--The nearest-neighbor heatmap appeared useful for helping identify outlier \textit{models}. Whenever the user loads a new model, the system automatically uses the new loaded model to find the nearest neighbor words to the query and updates the heatmap. When the user loads a model that results in significant disagreement with previously loaded models, there is a dramatic change on the heatmap which is very salient. As user P3 mentioned, ``When I turn the lockf up, I suddenly see the heatmap change all over, and this helps me know that something bad happens. So I am able to quickly recognize that and revert the change.''

\textit{Confirming Intuitions}--While all our participants were able to make use of the \lvt interface we found that those with high familiarity with embedding models used it in part to confirm intuitions. For example, P1, P2, and P7 believed the learning rate (from prior experience) should not be very big. By targeting \textit{alpha} they were able to obtain substantial improvement on both $f_T$ and $f_A$. Participants P2 and P5 mentioned that they knew \textit{window} cannot not be too big or too small and focused on tuning that parameter towards the middle region (4--8) and noted the gradual improvement on the performance (specially with $f_A$). Several participants in the \lvt group reported that being able to use the parallel coordinates plot helped them confirm such intuitions quickly.

\textit{Requested Features}--Multiple participants requested the ability to remove loaded models. This was surprising to us as we thought end-users would like to see models that did not work so they could avoid that area of the parameter space.  However, participants noted that these discarded models often introduced clutter especially in later stages of the workflow when they had narrowed down their search space and were focusing on debugging specific models (e.g., when they were using the query functionality). In the experiment, multiple parameters had non-linear effects on the performance. For example, the two extremes of \textit{lockf} and \textit{retro} had similar effects, while the middle-range values of the two parameters cause a linear change on the performance metric. P5 commented that this non-linear pattern was hard to identify with the parallel coordinates without any external knowledge. Augmented parallel coordinates, such as those with overlaid histograms, may alleviate this issue. The SPLOM view, which participants did not have access to, may also be useful.

\textit{Interface issues}--Most participants in our study were unfamiliar with parallel coordinates and found the brushing interactions to be unclear. We suspect that this is a common issue when learning parallel coordinates, but training and adding visual hints to the interface may help. P3 and P11 found the interactions of flagging word pairs as synonyms and/or antonyms to be cumbersome. Currently, to flag a word pair, the user has to find the target word in the overhead embedding explorer, and follows the connector to the corresponding column in the heatmap and click, and then in a popped-up window select the associated option. This interaction pattern can be simplified by allowing the user directly click on words in the embedding explorer to flag them.

%% file: 07_discussion.tex
\section{Discussion and Limitations}
\label{sec-lamvi:discuss}

We constructed \lvt for extension. However, we focused our implementation in a number of ways for the first iteration.  One specific issue is that only one `architecture' at a time is supported. There are many other embedding models including GloVe~\cite{pennington2014glove}, DeepWalk~\cite{perozzi2014deepwalk}, and LINE~\cite{tang2015line}. Not all utilize the same parameter space and an interesting question is how to allow the end-user to compare models which use very different parameters. A simple solution may be to allow parameters values to be ``null'' for models to which the parameter does not apply. Other, more complex, parallel coordinates configurations might also be an alternative~\cite{claessen11}. We also note that while we have adopted the dimensionality-reduction style visualization for the embeddings these sometimes lead to `tea leaf reading.' Alternative representations may yield more interpretable results~\cite{chuang12}.

\lvt is currently focused on \textit{word} embeddings. There are many other model types that might benefit from the exploratory features of \lvt, but various adaptations to the visualizations would be necessary. \lvt may be most readily adaptable for other embedding types (e.g., document/paragraph or bimodal~\cite{allamanis2015bimodal,le2014distributed}) as we can show other kinds of items in the t-SNE/PCA views. The workflow of inspecting vector similarity, vector interaction, and tracking the ranks of watched candidate items could be the same.  Other text-related architectures, such as those for memory networks~\cite{karpathy2015visualizing} that ``generate'' sentences would require other changes to the visualization panels. Similarly, image or audio related architectures would also require data-specific adaptations but we believe model comparison through the parallel coordinate views may still be useful in these contexts. In these cases the heatmap view may be adapted to show nearest neighbor images, salience maps, or pixel masks.

%% file: 08_conclusion.tex
\section{Conclusion}
\label{sec-lamvi:conclude}

In this paper we describe \lvt, a visual tool for comparing word embedding models. \lvt is specifically focused on three key decision points for machine learning model developers: (1) determining good hyperparameters for model testing, (2) find good examples for evaluations, and (3) selecting optimal ('satisficing') models. \lvt is designed to support these decisions through a focus on hyperparameters and their relationship to the model internals and performance metrics. \lvt combines different visual panels in an integrated view that supports these tasks. By extending a parallel coordinates display, \lvt allows the user to `pivot' between tasks to identify satisfactory hyperparameters. 

Unlike automated hyperparameter optimization algorithms, the interface allows the user to make decisions about how different metrics should be traded off. Additionally, the tool allows the end-user to isolate patterns in the embedding space and focus their attention on desirable models.  We demonstrate that \lvt can be used to gain insight into the relationship between hyperparameters.  Using a realistic experiment we also show how \lvt users can more rapidly identify these optimal configurations.

See \url{http://lamvi.info} for live demos.

%% file: 99_ack.tex
\section*{Acknowledgements}
We dedicate this paper to Xin Rong. It is based on an unpublished chapter from Xin Rong's doctoral dissertation. It is an important contribution that we felt should be shared.

%% file: 00_main.bbl
\begin{thebibliography}{10}
\renewcommand*{\sfdefault}{PTSansNarrow-TLF}

\bibitem{raytune}
Ray tune: Hyperparameter optimization framework.
\newblock \url{http://ray.readthedocs.io/en/latest/tune.html}.

\bibitem{alexander16}
E.~Alexander and M.~Gleicher.
\newblock Task-driven comparison of topic models.
\newblock {\em IEEE Transactions on Visualization and Computer Graphics},
  22(1):320--329, Jan 2016. doi: \textsf{%
10\hspace{.1pt}\discretionary{.}{%
}{.}\hspace{.4pt}1109\discretionary{/}{%
}{/}TVCG\hspace{.1pt}\discretionary{.}{%
}{.}\hspace{.4pt}2015\hspace{.1pt}\discretionary{.}{%
}{.}\hspace{.4pt}2467618}


\bibitem{allamanis2015bimodal}
M.~Allamanis, D.~Tarlow, A.~Gordon, and Y.~Wei.
\newblock Bimodal modelling of source code and natural language.
\newblock In {\em Proceedings of The $32^{nd}$ International Conference on
  Machine Learning}, pp. 2123--2132, 2015.

\bibitem{amershi2014power}
S.~Amershi, M.~Cakmak, W.~B. Knox, and T.~Kulesza.
\newblock Power to the people: The role of humans in interactive machine
  learning.
\newblock {\em AI Magazine}, 35(4):105--120, 2014.

\bibitem{modeltracker}
S.~Amershi, M.~Chickering, S.~M. Drucker, B.~Lee, P.~Simard, and J.~Suh.
\newblock Modeltracker: Redesigning performance analysis tools for machine
  learning.
\newblock In {\em Proceedings of the 33rd Annual ACM Conference on Human
  Factors in Computing Systems}, CHI '15, pp. 337--346. ACM, New York, NY, USA,
  2015. doi: \textsf{%
10\hspace{.1pt}\discretionary{.}{%
}{.}\hspace{.4pt}1145\discretionary{/}{%
}{/}2702123\hspace{.1pt}\discretionary{.}{%
}{.}\hspace{.4pt}2702509}


\bibitem{amershi10}
S.~Amershi, J.~Fogarty, A.~Kapoor, and D.~Tan.
\newblock Examining multiple potential models in end-user interactive concept
  learning.
\newblock In {\em Proceedings of the SIGCHI Conference on Human Factors in
  Computing Systems}, CHI '10, pp. 1357--1360. ACM, New York, NY, USA, 2010.
  doi: \textsf{%
10\hspace{.1pt}\discretionary{.}{%
}{.}\hspace{.4pt}1145\discretionary{/}{%
}{/}1753326\hspace{.1pt}\discretionary{.}{%
}{.}\hspace{.4pt}1753531}


\bibitem{rong2014word2vec}
Anonymized.
\newblock word2vec parameter learning explained.
\newblock {\em arXiv preprint arXiv:1411.2738}, 2014.

\bibitem{lamvidemo}
Anonymized.
\newblock Lamvi: Language model visual inspector [1.0].
\newblock \url{http://bit.ly/2xC7P8n}, 2016.

\bibitem{rong2016visual}
Anonymized.
\newblock Visual tools for debugging neural language models.
\newblock In {\em Proceedings of ICML Workshop on Visualization for Deep
  Learning}, 2016.

\bibitem{wevi}
Anonymized.
\newblock Wevi: Word embedding visual inspector.
\newblock \url{http://bit.ly/wevi-online}, 2016.

\bibitem{anthony2011antconc}
L.~Anthony.
\newblock Antconc (version 3.2. 2)[computer software].
\newblock {\em Tokyo, Japan: Waseda University}, 2011.

\bibitem{baccianella2010sentiwordnet}
S.~Baccianella, A.~Esuli, and F.~Sebastiani.
\newblock Sentiwordnet 3.0: An enhanced lexical resource for sentiment analysis
  and opinion mining.
\newblock In {\em LREC}, vol.~10, pp. 2200--2204, 2010.

\bibitem{bergstra2012random}
J.~Bergstra and Y.~Bengio.
\newblock Random search for hyper-parameter optimization.
\newblock {\em Journal of Machine Learning Research}, 13(Feb):281--305, 2012.

\bibitem{NIPS2011_4443}
J.~S. Bergstra, R.~Bardenet, Y.~Bengio, and B.~K\'{e}gl.
\newblock Algorithms for hyper-parameter optimization.
\newblock In J.~Shawe-Taylor, R.~S. Zemel, P.~L. Bartlett, F.~Pereira, and
  K.~Q. Weinberger, eds., {\em Advances in Neural Information Processing
  Systems 24}, pp. 2546--2554. Curran Associates, Inc., 2011.

\bibitem{bilal}
A.~Bilal, A.~Jourabloo, M.~Ye, X.~Liu, and L.~Ren.
\newblock Do convolutional neural networks learn class hierarchy?
\newblock {\em IEEE Transactions on Visualization and Computer Graphics},
  24(1):152--162, Jan 2018. doi: \textsf{%
10\hspace{.1pt}\discretionary{.}{%
}{.}\hspace{.4pt}1109\discretionary{/}{%
}{/}TVCG\hspace{.1pt}\discretionary{.}{%
}{.}\hspace{.4pt}2017\hspace{.1pt}\discretionary{.}{%
}{.}\hspace{.4pt}2744683}


\bibitem{bostock2011d3}
M.~Bostock, V.~Ogievetsky, and J.~Heer.
\newblock D$^3$ data-driven documents.
\newblock {\em IEEE transactions on visualization and computer graphics},
  17(12):2301--2309, 2011.

\bibitem{chuang2013topic}
J.~Chuang, S.~Gupta, C.~Manning, and J.~Heer.
\newblock Topic model diagnostics: Assessing domain relevance via topical
  alignment.
\newblock In {\em Proceedings of the 30th International Conference on Machine
  Learning (ICML-13)}, pp. 612--620, 2013.

\bibitem{chuang2012termite}
J.~Chuang, C.~D. Manning, and J.~Heer.
\newblock Termite: Visualization techniques for assessing textual topic models.
\newblock In {\em Proceedings of the International Working Conference on
  Advanced Visual Interfaces}, pp. 74--77. ACM, 2012.

\bibitem{chuang12}
J.~Chuang, D.~Ramage, C.~Manning, and J.~Heer.
\newblock Interpretation and trust: Designing model-driven visualizations for
  text analysis.
\newblock In {\em Proceedings of the SIGCHI Conference on Human Factors in
  Computing Systems}, CHI '12, pp. 443--452. ACM, New York, NY, USA, 2012. doi:
  \textsf{%
10\hspace{.1pt}\discretionary{.}{%
}{.}\hspace{.4pt}1145\discretionary{/}{%
}{/}2207676\hspace{.1pt}\discretionary{.}{%
}{.}\hspace{.4pt}2207738}


\bibitem{claessen11}
J.~H.~T. Claessen and J.~J. van Wijk.
\newblock Flexible linked axes for multivariate data visualization.
\newblock {\em IEEE Transactions on Visualization and Computer Graphics},
  17(12):2310--2316, Dec 2011. doi: \textsf{%
10\hspace{.1pt}\discretionary{.}{%
}{.}\hspace{.4pt}1109\discretionary{/}{%
}{/}TVCG\hspace{.1pt}\discretionary{.}{%
}{.}\hspace{.4pt}2011\hspace{.1pt}\discretionary{.}{%
}{.}\hspace{.4pt}201}


\bibitem{cui2014hierarchical}
W.~Cui, S.~Liu, Z.~Wu, and H.~Wei.
\newblock How hierarchical topics evolve in large text corpora.
\newblock {\em Visualization and Computer Graphics, IEEE Transactions on},
  20(12):2281--2290, 2014.

\bibitem{faruqui2014retrofitting}
M.~Faruqui, J.~Dodge, S.~K. Jauhar, C.~Dyer, E.~Hovy, and N.~A. Smith.
\newblock Retrofitting word vectors to semantic lexicons.
\newblock {\em arXiv preprint arXiv:1411.4166}, 2014.

\bibitem{ganitkevitch2013ppdb}
J.~Ganitkevitch, B.~Van~Durme, and C.~Callison-Burch.
\newblock Ppdb: The paraphrase database.
\newblock In {\em HLT-NAACL}, pp. 758--764, 2013.

\bibitem{vizier}
D.~Golovin, B.~Solnik, S.~Moitra, G.~Kochanski, J.~Karro, and D.~Sculley.
\newblock Google vizier: A service for black-box optimization.
\newblock In {\em Proceedings of the 23rd ACM SIGKDD International Conference
  on Knowledge Discovery and Data Mining}, KDD '17, pp. 1487--1495. ACM, New
  York, NY, USA, 2017. doi: \textsf{%
10\hspace{.1pt}\discretionary{.}{%
}{.}\hspace{.4pt}1145\discretionary{/}{%
}{/}3097983\hspace{.1pt}\discretionary{.}{%
}{.}\hspace{.4pt}3098043}


\bibitem{greff2016lstm}
K.~Greff, R.~K. Srivastava, J.~Koutn{\'\i}k, B.~R. Steunebrink, and
  J.~Schmidhuber.
\newblock Lstm: A search space odyssey.
\newblock {\em IEEE transactions on neural networks and learning systems},
  2016.

\bibitem{hohman2018visual}
F.~Hohman, M.~Kahng, R.~Pienta, and D.~H. Chau.
\newblock Visual analytics in deep learning: An interrogative survey for the
  next frontiers.
\newblock {\em arXiv preprint arXiv:1801.06889}, 2018.

\bibitem{inselberg1985plane}
A.~Inselberg.
\newblock The plane with parallel coordinates.
\newblock {\em The visual computer}, 1(2):69--91, 1985.

\bibitem{detective}
A.~Inselberg.
\newblock Multidimensional detective.
\newblock In {\em Information Visualization, 1997. Proceedings., IEEE Symposium
  on}, pp. 100--107, Oct 1997. doi: \textsf{%
10\hspace{.1pt}\discretionary{.}{%
}{.}\hspace{.4pt}1109\discretionary{/}{%
}{/}INFVIS\hspace{.1pt}\discretionary{.}{%
}{.}\hspace{.4pt}1997\hspace{.1pt}\discretionary{.}{%
}{.}\hspace{.4pt}636793}


\bibitem{jolliffe2002principal}
I.~Jolliffe.
\newblock {\em Principal component analysis}.
\newblock Wiley Online Library, 2002.

\bibitem{semeval}
D.~A. Jurgens, P.~D. Turney, S.~M. Mohammad, and K.~J. Holyoak.
\newblock Semeval-2012 task 2: Measuring degrees of relational similarity.
\newblock In {\em Proceedings of the First Joint Conference on Lexical and
  Computational Semantics - Volume 1: Proceedings of the Main Conference and
  the Shared Task, and Volume 2: Proceedings of the Sixth International
  Workshop on Semantic Evaluation}, SemEval '12, pp. 356--364. Association for
  Computational Linguistics, Stroudsburg, PA, USA, 2012.

\bibitem{activis}
M.~Kahng, P.~Y. Andrews, A.~Kalro, and D.~H.~. Chau.
\newblock Activis: Visual exploration of industry-scale deep neural network
  models.
\newblock {\em IEEE Transactions on Visualization and Computer Graphics},
  24(1):88--97, Jan 2018. doi: \textsf{%
10\hspace{.1pt}\discretionary{.}{%
}{.}\hspace{.4pt}1109\discretionary{/}{%
}{/}TVCG\hspace{.1pt}\discretionary{.}{%
}{.}\hspace{.4pt}2017\hspace{.1pt}\discretionary{.}{%
}{.}\hspace{.4pt}2744718}


\bibitem{datacube}
M.~Kahng, D.~Fang, and D.~H.~P. Chau.
\newblock Visual exploration of machine learning results using data cube
  analysis.
\newblock In {\em Proceedings of the Workshop on Human-In-the-Loop Data
  Analytics}, HILDA '16, pp. 1:1--1:6. ACM, New York, NY, USA, 2016. doi:
  \textsf{%
10\hspace{.1pt}\discretionary{.}{%
}{.}\hspace{.4pt}1145\discretionary{/}{%
}{/}2939502\hspace{.1pt}\discretionary{.}{%
}{.}\hspace{.4pt}2939503}


\bibitem{karpathy2015visualizing}
A.~Karpathy, J.~Johnson, and F.-F. Li.
\newblock Visualizing and understanding recurrent networks.
\newblock {\em arXiv preprint arXiv:1506.02078}, 2015.

\bibitem{kruskal1964multidimensional}
J.~B. Kruskal.
\newblock Multidimensional scaling by optimizing goodness of fit to a nonmetric
  hypothesis.
\newblock {\em Psychometrika}, 29(1):1--27, 1964.

\bibitem{Kulesza:2015:PED:2678025.2701399}
T.~Kulesza, M.~Burnett, W.-K. Wong, and S.~Stumpf.
\newblock Principles of explanatory debugging to personalize interactive
  machine learning.
\newblock In {\em Proceedings of the 20th International Conference on
  Intelligent User Interfaces}, IUI '15, pp. 126--137. ACM, New York, NY, USA,
  2015. doi: \textsf{%
10\hspace{.1pt}\discretionary{.}{%
}{.}\hspace{.4pt}1145\discretionary{/}{%
}{/}2678025\hspace{.1pt}\discretionary{.}{%
}{.}\hspace{.4pt}2701399}


\bibitem{kulesza2011oriented}
T.~Kulesza, S.~Stumpf, W.-K. Wong, M.~M. Burnett, S.~Perona, A.~Ko, and
  I.~Oberst.
\newblock Why-oriented end-user debugging of naive bayes text classification.
\newblock {\em ACM Transactions on Interactive Intelligent Systems (TiiS)},
  1(1):2, 2011.

\bibitem{le2014distributed}
Q.~V. Le and T.~Mikolov.
\newblock Distributed representations of sentences and documents.
\newblock {\em arXiv preprint arXiv:1405.4053}, 2014.

\bibitem{lindberg2012oxford}
C.~A. Lindberg.
\newblock {\em Oxford American writer's thesaurus}.
\newblock Oxford University Press, USA, 2012.

\bibitem{liushi}
M.~Liu, J.~Shi, K.~Cao, J.~Zhu, and S.~Liu.
\newblock Analyzing the training processes of deep generative models.
\newblock {\em IEEE Transactions on Visualization and Computer Graphics},
  24(1):77--87, Jan 2018. doi: \textsf{%
10\hspace{.1pt}\discretionary{.}{%
}{.}\hspace{.4pt}1109\discretionary{/}{%
}{/}TVCG\hspace{.1pt}\discretionary{.}{%
}{.}\hspace{.4pt}2017\hspace{.1pt}\discretionary{.}{%
}{.}\hspace{.4pt}2744938}


\bibitem{LIU201748}
S.~Liu, X.~Wang, M.~Liu, and J.~Zhu.
\newblock Towards better analysis of machine learning models: A visual
  analytics perspective.
\newblock {\em Visual Informatics}, 1(1):48 -- 56, 2017. doi: \textsf{%
10\hspace{.1pt}\discretionary{.}{%
}{.}\hspace{.4pt}1016\discretionary{/}{%
}{/}j\hspace{.1pt}\discretionary{.}{%
}{.}\hspace{.4pt}visinf\hspace{.1pt}\discretionary{.}{%
}{.}\hspace{.4pt}2017\hspace{.1pt}\discretionary{.}{%
}{.}\hspace{.4pt}01\hspace{.1pt}\discretionary{.}{%
}{.}\hspace{.4pt}006}


\bibitem{liu2009interactive}
S.~Liu, M.~X. Zhou, S.~Pan, W.~Qian, W.~Cai, and X.~Lian.
\newblock Interactive, topic-based visual text summarization and analysis.
\newblock In {\em Proceedings of the 18th ACM conference on Information and
  knowledge management}, pp. 543--552. ACM, 2009.

\bibitem{mikolov2007language}
T.~Mikolov.
\newblock {\em Language Modeling for Speech Recognition in Czech}.
\newblock PhD thesis, Masters thesis, Brno University of Technology, 2007.

\bibitem{mikolov2013efficient}
T.~Mikolov, K.~Chen, G.~Corrado, and J.~Dean.
\newblock Efficient estimation of word representations in vector space.
\newblock {\em arXiv preprint arXiv:1301.3781}, 2013.

\bibitem{mikolov2013distributed}
T.~Mikolov, I.~Sutskever, K.~Chen, G.~S. Corrado, and J.~Dean.
\newblock {Distributed Representations of Words and Phrases and their
  Compositionality}.
\newblock In {\em Advances in Neural Information Processing Systems 26}, pp.
  3111--3119. Curran Associates, Inc., 2013.

\bibitem{ming1understanding}
Y.~Ming, S.~Cao, R.~Zhang, Z.~Li, and Y.~Chen.
\newblock Understanding hidden memories of recurrent neural networks.
\newblock {\em dim}, 1(1.0):0--5.

\bibitem{munzner2014visualization}
T.~Munzner.
\newblock {\em Visualization analysis and design}.
\newblock CRC Press, 2014.

\bibitem{olah2018the}
C.~Olah, A.~Satyanarayan, I.~Johnson, S.~Carter, L.~Schubert, K.~Ye, and
  A.~Mordvintsev.
\newblock The building blocks of interpretability.
\newblock {\em Distill}, 2018.
\newblock https://distill.pub/2018/building-blocks. doi: \textsf{%
10\hspace{.1pt}\discretionary{.}{%
}{.}\hspace{.4pt}23915\discretionary{/}{%
}{/}distill\hspace{.1pt}\discretionary{.}{%
}{.}\hspace{.4pt}00010}


\bibitem{pang2002thumbs}
B.~Pang, L.~Lee, and S.~Vaithyanathan.
\newblock Thumbs up?: sentiment classification using machine learning
  techniques.
\newblock In {\em Proceedings of the ACL-02 conference on Empirical methods in
  natural language processing-Volume 10}, pp. 79--86. Association for
  Computational Linguistics, 2002.

\bibitem{conceptvector}
D.~Park, S.~Kim, J.~Lee, J.~Choo, N.~Diakopoulos, and N.~Elmqvist.
\newblock Conceptvector: Text visual analytics via interactive lexicon building
  using word embedding.
\newblock {\em IEEE Transactions on Visualization and Computer Graphics},
  PP(99):1--1, 2017. doi: \textsf{%
10\hspace{.1pt}\discretionary{.}{%
}{.}\hspace{.4pt}1109\discretionary{/}{%
}{/}TVCG\hspace{.1pt}\discretionary{.}{%
}{.}\hspace{.4pt}2017\hspace{.1pt}\discretionary{.}{%
}{.}\hspace{.4pt}2744478}


\bibitem{Patel:2008:ISM:1357054.1357160}
K.~Patel, J.~Fogarty, J.~A. Landay, and B.~Harrison.
\newblock Investigating statistical machine learning as a tool for software
  development.
\newblock In {\em Proceedings of the SIGCHI Conference on Human Factors in
  Computing Systems}, CHI '08, pp. 667--676. ACM, New York, NY, USA, 2008. doi:
  \textsf{%
10\hspace{.1pt}\discretionary{.}{%
}{.}\hspace{.4pt}1145\discretionary{/}{%
}{/}1357054\hspace{.1pt}\discretionary{.}{%
}{.}\hspace{.4pt}1357160}


\bibitem{pennington2014glove}
J.~Pennington, R.~Socher, and C.~D. Manning.
\newblock Glove: Global vectors for word representation.
\newblock In {\em EMNLP}, vol.~14, pp. 1532--1543, 2014.

\bibitem{perozzi2014deepwalk}
B.~Perozzi, R.~Al-Rfou, and S.~Skiena.
\newblock Deepwalk: Online learning of social representations.
\newblock In {\em Proceedings of the 20th ACM SIGKDD international conference
  on Knowledge discovery and data mining}, pp. 701--710. ACM, 2014.

\bibitem{deepeyes}
N.~Pezzotti, T.~Höllt, J.~V. Gemert, B.~P.~F. Lelieveldt, E.~Eisemann, and
  A.~Vilanova.
\newblock Deepeyes: Progressive visual analytics for designing deep neural
  networks.
\newblock {\em IEEE Transactions on Visualization and Computer Graphics},
  24(1):98--108, Jan 2018. doi: \textsf{%
10\hspace{.1pt}\discretionary{.}{%
}{.}\hspace{.4pt}1109\discretionary{/}{%
}{/}TVCG\hspace{.1pt}\discretionary{.}{%
}{.}\hspace{.4pt}2017\hspace{.1pt}\discretionary{.}{%
}{.}\hspace{.4pt}2744358}


\bibitem{rao1994table}
R.~Rao and S.~K. Card.
\newblock The table lens: merging graphical and symbolic representations in an
  interactive focus+ context visualization for tabular information.
\newblock In {\em Proceedings of the SIGCHI conference on Human factors in
  computing systems}, pp. 318--322. ACM, 1994.

\bibitem{rehurek}
R.~Rehurek.
\newblock Deep learning with word2vec.
\newblock \url{https://radimrehurek.com/gensim/models/word2vec.html}, 2013.

\bibitem{samek2015evaluating}
W.~Samek, A.~Binder, G.~Montavon, S.~Bach, and K.-R. M{\"u}ller.
\newblock Evaluating the visualization of what a deep neural network has
  learned.
\newblock {\em arXiv preprint arXiv:1509.06321}, 2015.

\bibitem{sievert2014ldavis}
C.~Sievert and K.~E. Shirley.
\newblock Ldavis: A method for visualizing and interpreting topics.
\newblock In {\em Proceedings of the workshop on interactive language learning,
  visualization, and interfaces}, pp. 63--70, 2014.

\bibitem{DBLP:journals/corr/SimonyanVZ13}
K.~Simonyan, A.~Vedaldi, and A.~Zisserman.
\newblock Deep inside convolutional networks: Visualising image classification
  models and saliency maps.
\newblock {\em CoRR}, abs/1312.6034, 2013.

\bibitem{NIPS2012_4522}
J.~Snoek, H.~Larochelle, and R.~P. Adams.
\newblock Practical bayesian optimization of machine learning algorithms.
\newblock In F.~Pereira, C.~J.~C. Burges, L.~Bottou, and K.~Q. Weinberger,
  eds., {\em Advances in Neural Information Processing Systems 25}, pp.
  2951--2959. Curran Associates, Inc., 2012.

\bibitem{DBLP:journals/corr/StrobeltGHPR16}
H.~Strobelt, S.~Gehrmann, B.~Huber, H.~Pfister, and A.~M. Rush.
\newblock Visual analysis of hidden state dynamics in recurrent neural
  networks.
\newblock {\em CoRR}, abs/1606.07461, 2016.

\bibitem{tang2015line}
J.~Tang, M.~Qu, M.~Wang, M.~Zhang, J.~Yan, and Q.~Mei.
\newblock Line: Large-scale information network embedding.
\newblock In {\em Proceedings of the 24th International Conference on World
  Wide Web}, pp. 1067--1077. International World Wide Web Conferences Steering
  Committee, 2015.

\bibitem{van2008visualizing}
L.~Van~der Maaten and G.~Hinton.
\newblock Visualizing data using t-sne.
\newblock {\em Journal of Machine Learning Research}, 9(2579-2605):85, 2008.

\bibitem{wattenberg2016how}
M.~Wattenberg, F.~Viégas, and I.~Johnson.
\newblock How to use t-sne effectively.
\newblock {\em Distill}, 2016. doi: \textsf{%
10\hspace{.1pt}\discretionary{.}{%
}{.}\hspace{.4pt}23915\discretionary{/}{%
}{/}distill\hspace{.1pt}\discretionary{.}{%
}{.}\hspace{.4pt}00002}


\bibitem{8019861}
K.~Wongsuphasawat, D.~Smilkov, J.~Wexler, J.~Wilson, D.~Mané, D.~Fritz,
  D.~Krishnan, F.~B. Viégas, and M.~Wattenberg.
\newblock Visualizing dataflow graphs of deep learning models in tensorflow.
\newblock {\em IEEE Transactions on Visualization and Computer Graphics},
  24(1):1--12, Jan 2018. doi: \textsf{%
10\hspace{.1pt}\discretionary{.}{%
}{.}\hspace{.4pt}1109\discretionary{/}{%
}{/}TVCG\hspace{.1pt}\discretionary{.}{%
}{.}\hspace{.4pt}2017\hspace{.1pt}\discretionary{.}{%
}{.}\hspace{.4pt}2744878}


\bibitem{yosinski2015understanding}
J.~Yosinski, J.~Clune, A.~Nguyen, T.~Fuchs, and H.~Lipson.
\newblock Understanding neural networks through deep visualization.
\newblock {\em arXiv preprint arXiv:1506.06579}, 2015.

\bibitem{Young2015}
S.~R. Young, D.~C. Rose, T.~P. Karnowski, S.-H. Lim, and R.~M. Patton.
\newblock Optimizing deep learning hyper-parameters through an evolutionary
  algorithm.
\newblock In {\em Proceedings of the Workshop on Machine Learning in
  High-Performance Computing Environments}, MLHPC '15, pp. 4:1--4:5. ACM, New
  York, NY, USA, 2015. doi: \textsf{%
10\hspace{.1pt}\discretionary{.}{%
}{.}\hspace{.4pt}1145\discretionary{/}{%
}{/}2834892\hspace{.1pt}\discretionary{.}{%
}{.}\hspace{.4pt}2834896}


\bibitem{DBLP:journals/corr/ZeilerF13}
M.~D. Zeiler and R.~Fergus.
\newblock Visualizing and understanding convolutional networks.
\newblock {\em CoRR}, abs/1311.2901, 2013.

\end{thebibliography}
